\documentclass[11pt]{article}

\usepackage[preprint]{acl}

\usepackage{times}
\usepackage{latexsym}

\usepackage[T1]{fontenc}
\usepackage[utf8]{inputenc}

\usepackage{microtype}

\usepackage{inconsolata}

\usepackage{graphicx}

\usepackage{booktabs,multirow}
\usepackage{amssymb}
\usepackage{float}

\usepackage{booktabs,tabularx,siunitx,makecell,graphicx,xcolor,colortbl,pifont}
\usepackage{url} % or \usepackage[hidelinks]{hyperref}
\usepackage{adjustbox}

\usepackage{subfig}

\newcommand{\yes}{\textcolor{green!60!black}{\ding{51}}}
\newcommand{\no}{\textcolor{red!70!black}{\ding{55}}}

\newcommand{\benchname}{AoT-PsyPhyBENCH}
\definecolor{orange}{RGB}{255,165,0}
\usepackage{listings}
\usepackage{xcolor}
\definecolor{promptbg}{RGB}{248,248,248}
\definecolor{promptframe}{RGB}{200,200,200}

\definecolor{darkgreen}{HTML}{1B5E20} % deep green
\definecolor{darkred}{HTML}{B71C1C}   % deep red

\lstdefinestyle{promptbox}{
  basicstyle=\footnotesize\sffamily,
  backgroundcolor=\color{promptbg},
  frame=single,
  rulecolor=\color{promptframe},
  breaklines=true,
  breakautoindent=false,   % <-- no hanging indent on wrapped lines
  breakindent=0pt,         % <-- zero indent for wrapped lines
  showstringspaces=false,
  columns=fullflexible,
  keepspaces=true,
  aboveskip=0.25\baselineskip,
  belowskip=0.25\baselineskip,
  framesep=4pt,
  xleftmargin=0pt,
  xrightmargin=0pt,
  framexleftmargin=0pt,
  escapeinside={(*@}{@*)},
  moredelim=**[is][\color{darkgreen}]{<<green>>}{<</green>>},
  moredelim=**[is][\color{darkred}]{<<red>>}{<</red>>},
  moredelim=**[is][\color{orange}]{<<orange>>}{<</orange>>},
  moredelim=**[is][\bfseries]{<<bf>>}{<</bf>>},
}

\usepackage{graphicx,calc}
\newsavebox{\PromptBox}

\newlength{\LeftW}   % width of the left image column
\newlength{\GapW}    % gap between columns

\title{Which Way Does Time Flow? A Psychophysics-Grounded Evaluation for Vision–Language Models}

\author{
Shiho Matta\textsuperscript{1,*},
Lis K. Pereira\textsuperscript{2,3,4,*},
Peitao Han\textsuperscript{2,3,4},
Shigeru Kitazawa\textsuperscript{2,3,4},
Fei Cheng\textsuperscript{1} \\
\textsuperscript{1} Kyoto University, Japan \\
\textsuperscript{2} Center for Information and Neural Networks, Japan \\
\textsuperscript{3} National Institute of Information and Communications Technology, Japan \\
\textsuperscript{4} The University of Osaka, Japan \\
\textsuperscript{*} \texttt{matta@nlp.ist.i.kyoto-u.ac.jp},
\texttt{liskanashiro@nict.go.jp}
}

\begin{document}
\maketitle
\begin{abstract}
Modern vision–language models (VLMs) excel at many multimodal tasks, yet their grasp of temporal information in video remains weak and has not been adequately evaluated. We probe this gap with a deceptively simple but revealing challenge: judging the arrow of time (AoT)—whether a short clip is played forward or backward. We introduce \textbf{\benchname}, a psychophysically validated benchmark that tests whether VLMs can infer temporal direction in natural videos using the same stimuli and behavioral baselines established for humans. Our comprehensive evaluation of open-weight and proprietary, reasoning and non-reasoning VLMs reveals that most models perform near chance, and even the best model lags far behind human accuracy on physically irreversible processes (e.g., free fall, diffusion/explosion) and causal manual actions (division/addition) that humans recognize almost instantly. These results highlight a fundamental gap in current multimodal systems: while they capture rich visual–semantic correlations, they lack the inductive biases required for temporal continuity and causal understanding. 
We release the code and data for {\benchname} to encourage further progress in the physical and temporal reasoning capabilities of VLMs.
% \benchname~ provides a rigorous, human-grounded diagnostic for advancing physical and temporal reasoning in VLMs.
\end{abstract}
\section{Introduction}

Vision–language models (VLMs) have recently achieved remarkable progress in multimodal understanding, including tasks such as video captioning, retrieval, and question answering \cite{bai2025qwen2, qwen_qvq_2024, liu-etal-2024-mibench, wang2024muirbench, xiao2021nextqa,wu2024starbench,fu2025videomme}. 
%Yet a fundamental question remains: do these models possess the inductive biases that humans rely on to interpret the physical world? 
Yet a fundamental question remains: do these models possess the implicit physical and causal assumptions, i.e. the \emph{inductive biases}, that humans rely on to interpret the physical world? 
Among the most basic of such biases is the \emph{arrow of time} (AoT): the implicit assumption that events unfold irreversibly from past to future, constrained by gravity, entropy, and causality. Judging whether a short video plays forward or backward provides a minimal yet powerful diagnostic of such bias: a system that has internalized temporal irreversibility should recognize when physical regularities are violated, whereas a system that relies only on visual correlation will fail despite extensive training.

\begin{figure}[!th]
    \centering
    \includegraphics[width=\columnwidth]{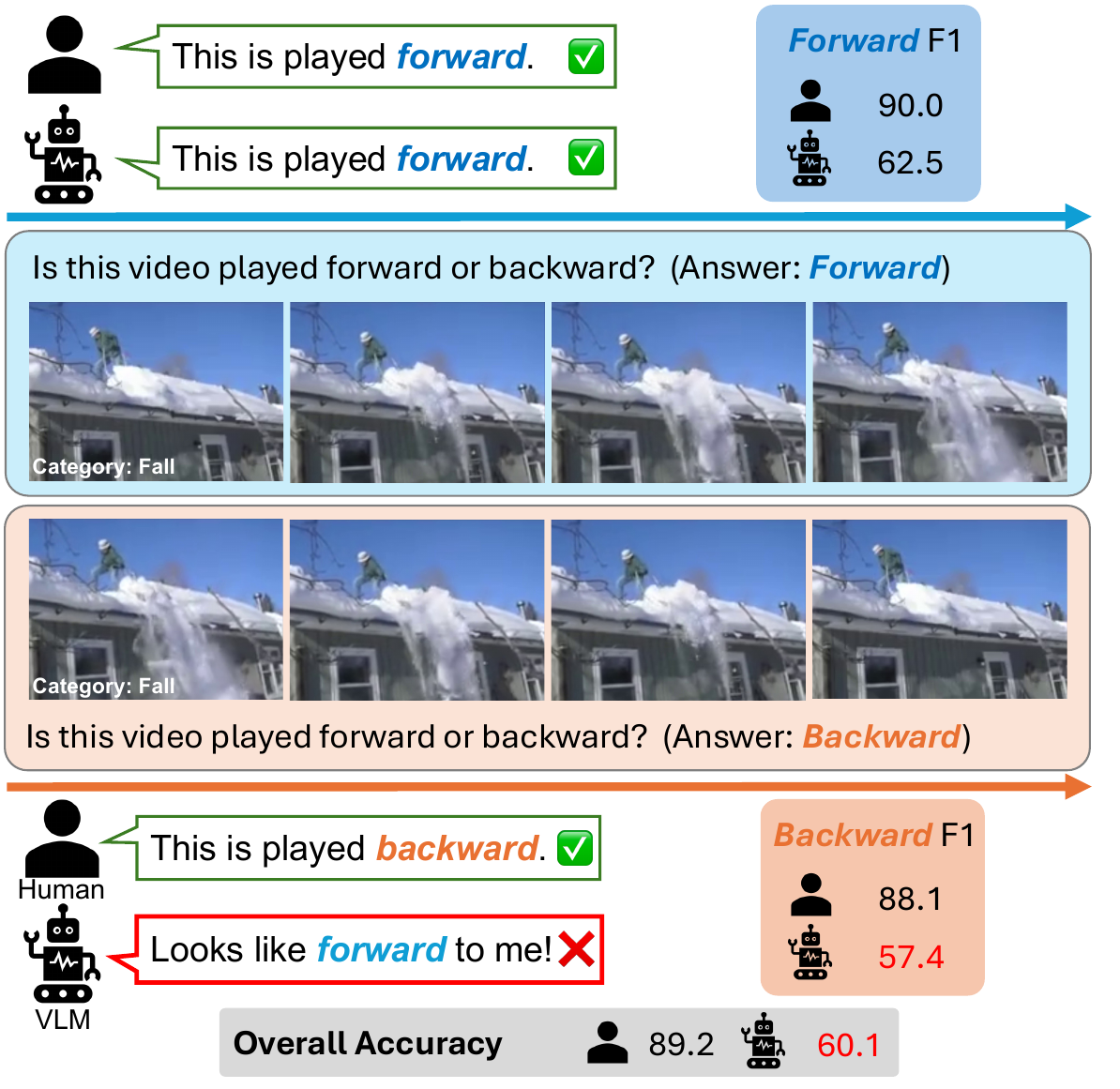}
    \caption{Overview of the arrow-of-time (AoT) task. Given a video played either forward or backward, the model must predict its temporal direction. Humans identify temporal direction accurately, whereas VLMs perform worse and exhibit a bias toward Forward (Section~\ref{sec:label_prediction_bias}). Human Forward/Backward F1 is computed on AoT-PsyPhyBENCH, excluding the Reciprocal (cyclic-motion) category.}
    \label{fig:task_overview}
\end{figure}
To anchor this diagnostic in human performance, \citet{hanyu2023ready} conducted a psychophysical study using 360 three-second natural video clips from everyday activities. They found a global forward bias: participants more often judged reversed clips as forward (39\% errors on reversed vs. 9\% on forward trials). However, humans detected reversals rapidly and almost flawlessly in five specific categories: \textbf{Fall} (free fall) and \textbf{Diffusion} (diffusion/explosion), which involve physically irreversible processes governed by gravity and entropy; \textbf{Proceed} (forward locomotion); and \textbf{Division} (manual division) and \textbf{Put} (manual addition), which involve agent-driven causal sequences.
%(1) free fall and (2) diffusion/explosion, which involve physically irreversible processes governed by gravity and entropy; (3) forward locomotion; and (4) manual division and (5) manual addition, which involve agent-driven causal sequences.
In these cases, reversed playback becomes visibly implausible. For example, when a big chunk of snow ascends back onto the roof, defying gravity (Figure~\ref{fig:task_overview}). These results demonstrate that humans exploit not only low-level motion cues but prior knowledge of physical regularities and causal structure, providing a psychophysically validated foundation for testing AoT in VLMs.

Building on this, we introduce \textbf{\benchname} (pronounced “\textit{AoT-sci-fi bench}”), an evaluation framework that directly inherits validated stimuli and behavioral baselines from \citet{hanyu2023ready}. We systematically test modern VLMs, from open-weight to proprietary reasoning/non-reasoning models, on the same AoT judgment task. Our findings reveal that most models perform near chance even in "obvious" irreversible processes, with the best setting lagging humans by approximately \textbf{29 percentage points}, suggesting that they lack the inductive biases underpinning human causal and temporal perception (Figure \ref{fig:task_overview}).

\textbf{Our contributions are threefold.}
\textbf{(1) Comprehensive evaluation.} We present the first systematic evaluation of modern VLMs on psychophysically grounded AoT judgment across zero-shot, few-shot, chain-of-thought prompting, and supervised fine-tuning. Despite apparent task simplicity, the gains from elaborate prompting and fine-tuning are minimal and inconsistent, indicating that the bottleneck is temporal and physical understanding, not instruction following.
\textbf{(2) Psychophysically validated benchmark.} We introduce \textbf{\benchname}, curated from \citeauthor{hanyu2023ready}'s psychophysical dataset by retaining only clips with clear human consensus on irreversible processes while excluding ambiguous cyclic motions. This yields a low-ambiguity benchmark enabling direct, reliable human–model comparison on temporal direction judgment.
\textbf{(3) Open resources.} We release our benchmark, evaluation scripts, and model outputs to advance research on temporal reasoning and physical understanding in VLMs.\footnote{\url{https://mattashiho233.github.io/AoT-PsyPhyBENCH-website/}}

\section{Related Work}

\label{subsec:vlms}
\subsection{Vision–Language Models (VLMs)}
%Modern VLMs combine a vision encoder with a large language model (LLM) through a multimodal connector, typically aligned via instruction tuning on image and/or video-text data. For video understanding, VLMs typically sample at a low frame rate or select a fixed numbers of frames, treating them as separate images with orders. Commonly, each frame is first encoded using a pre-trianed image encoder (e.g.: (e.g., ViT), then mapped to the text space via a modality aligner, and finally fed into an LLM backbone for response generation. Although efficient, low rate frame sampling can lead to critical visual information loss. A consequence is that, even with frontier VLMs claiming they have achieved deep video understading capabilities, they still fail on tasks that depend on event order.

Modern VLMs couple a vision encoder with a large language model (LLM) via a multimodal connector and are typically aligned through instruction tuning on image–/video–text data \cite{wang2024qwen2vl, qwen_qvq_2024, azzolini2025cosmos, bai2025qwen2}. %For video, most pipelines sample at a low frame rate per second (usually $FPS \leq 2$) or select a fixed number of keyframes, then treat frames largely as independent images with shallow order cues. In a common design, each frame is encoded by a pre-trained image encoder (e.g. ViT, CLIP), projected into the language space by a modality aligner, and passed to an LLM for reasoning and generation. While computationally efficient, low-rate sampling and image-only encoders discard high-frequency motion and contact dynamics, introducing temporal aliasing and information loss. As a result, even frontier VLMs that claim deep video understanding often fail on tasks that depend on event order (e.g. AoT judgements) \cite{xue2025seeing}.
Despite rapid progress, frontier VLMs often fail on video tasks that depend on event order  \cite{xue2025seeing}.

%Despite rapid advances, frontier VLMs continue to \emph{struggle} with AoT judgements, i.e., deciding whether a clip is \texttt{forward} or \texttt{reversed}.

We broadly categorized the VLMs used in our evaluation along two axes: \textbf{(i) proprietary vs. open-weight} and \textbf{(ii) reasoning vs. non-reasoning}. Proprietary models are accessible only via APIs with limited transparency, while open-weight models release checkpoints and inference scripts, as well as partial training details. Reasoning models are trained to generate multi-step deliberation (e.g., chain-of-thought reasoning) automatically before producing the final output, often with controllable \emph{reasoning effort}.
Non-reasoning models prioritise perceptual understanding (e.g., captioning, visual question-answering) and instruction following, without dedicated reasoning objectives; they typically respond directly unless explicitly prompted for step-by-step reasoning.

Building on this taxonomy, we evaluate the following models. \textbf{Proprietary non-reasoning:} GPT-4o and GPT-4.1~\cite{openai_gpt4o_blog_2024,openai_gpt41_blog_2025}. \textbf{Proprietary reasoning:} o3, o4-mini, and GPT-5~\cite{openai_o3_blog_2025,openai_gpt5_blog_2025}; Gemini-2.5-Pro~\cite{comanici2025gemini}. \textbf{Open-weight non-reasoning:} Qwen2-VL~\cite{wang2024qwen2vl} and Qwen2.5-VL~\cite{bai2025qwen2}. \textbf{Open-weight reasoning:} Cosmos-Reason1~\cite{azzolini2025cosmos}, which combines explicit AoT supervision and reinforcement learning on $\sim$30k pairs of forward/reverse video clips with reasoning traces; and QVQ-72B-Preview~\cite{qwen_qvq_2024}.

\subsection{Temporal Reasoning Benchmarks}

We evaluate VLMs on a psychophysically validated AoT benchmark to address two key limitations in current temporal reasoning evaluation:

\textbf{(1) Lack of temporal dependency.} Several recent benchmarks claim to assess temporal reasoning in VLMs, but many do not \emph{demonstrate} dependence on event order \cite{liu-etal-2024-mibench, wang2024muirbench, fu2025videomme, xiao2021nextqa, wu2024starbench}. \citet{xue2025seeing} evaluated strong VLMs on standard temporal benchmarks and found a critical failure: models show minimal or no performance degradation when frames are \emph{shuffled} or \emph{reversed}, indicating these tasks can be solved through scene context without temporal understanding. In contrast, the AoT judgment is inherently temporal: reversed videos violate physical laws, making temporal order task critical, rather than incidental.

\textbf{(2) Lack of psychophysical validation.} Existing AoT benchmarks simply reverse video datasets~\cite{bagad2023test,wang2023paxion,du2024reversed,agarwal2025cosmos,xue2025seeing} without \emph{controlled human baselines}, for instance, with quantitative measurements of accuracy, reaction times, and category-specific effects. This creates systematic evaluation problems: difficulty remains uncalibrated (videos ambiguous to humans become spurious failure cases), model–human performance gaps cannot be quantified, systematic biases (forward bias, category effects) go undetected, and bidirectionally plausible cases contaminate test sets. Psychophysical grounding is essential to distinguish genuine deficiencies from artifacts of ambiguous stimuli.

To address these limitations, we introduce \textbf{\benchname}, a psychophysically validated benchmark enabling direct human–model comparison on temporal direction judgment.

\section{\benchname: A Psychophysically Validated Arrow-of-Time Benchmark}

% We evaluate whether vision–language models (VLMs) can infer the arrow of time—that is, discriminate \texttt{forward} from \texttt{reverse} playback—in short, everyday videos.
We evaluate whether vision--language models (VLMs) can infer the arrow of time, that is, distinguish between \textit{forward} and \textit{backward} playbacks in everyday videos. Our benchmark builds on the psychophysical study of \citet{hanyu2023ready}, enabling direct comparison with human performance. The source dataset comprises 360 three-second clips (29.97 FPS) from the Moments in Time dataset \cite{monfort2019moments}, covering a broad range of daily dynamics.
%: one clip from each of 339 action classes plus 21 additional clips sampled from randomly selected classes, covering a broad range of daily dynamics.
Ten participants viewed every clip twice—once forward and once reversed—across two sessions on different days (each session: 180 forward, 180 reversed; order randomized per participant). Three raters annotated six motion categories: (1) \textbf{Proceed}: forward locomotion of people, animals, or vehicles; (2) \textbf{Fall}: free-fall/ballistic motion under gravity; (3) \textbf{Diffusion}: centrifugal diffusion or small-particle explosions; (4) \textbf{Division}: division of material by hand or tool; (5) \textbf{Put}: addition/construction of material by hand; and (6) \textbf{Reciprocal}: reciprocating (cyclic) motion (Table~\ref{tab:motion_categories_figures}). The first five categories reflect \emph{irreversible} processes that, when reversed, tend to violate fundamental physical regularities (entropy increase, gravity, causal ordering), whereas reciprocating motion is \emph{bidirectional} and often appears plausible in both directions. Categories are not mutually exclusive.
Human performance exhibited a pronounced \emph{forward bias}: participants made errors on \textbf{39\%} of reversed trials versus \textbf{9\%} of forward trials, with accuracy strongly modulated by motions from the first five categories. Performance was significantly lower for the reciprocating motion category.

We introduce \textbf{\benchname}, a curated subset of \citet{hanyu2023ready} that \emph{excludes the reciprocal motion category} (category 6) and retains only high-consensus clips from categories 1–5. This exclusion is justified by human performance: the reciprocal motion category achieved only 61.1\% overall accuracy (accuracy with forward and backward videos combined) compared to 81.3–85.1\% for the other motion categories, substantially degrading benchmark reliability (77.6\% with cyclic motion vs. 89.0\% without). From the original 360 clips (720 videos with reversals), our filtering retains 212 clips, resulting in a total of \textbf{424} videos.
%From the original 720 videos (360 clips with their reversed counterparts), our filtering yields 212 clips (totaling \textbf{424} videos with reversals). 
This yields a low-ambiguity benchmark that enables direct and reliable comparison between humans and models. Table~\ref{tab:motion_categories_figures} provides an overview of the motion categories with comparative statistics illustrating how \benchname~ filters the original dataset. 

% In this work, we introduce \textbf{\benchname}, a curated subset of \citet{hanyu2023ready} that \emph{excludes the reciprocating (cyclic) motion category} (category 6) and retains only high-consensus clips from categories 1–5. This exclusion is justified by human performance data: the cyclic motion category achieved only 61.1\% overall accuracy (accuracy with forward and backward videos combined) compared to 83.8–93.5\% for the other motion categories, substantially degrading overall benchmark reliability (77.6\% with cyclic motion vs. 89.2\% without). From the original 720 videos (360 clips with their reversed counterparts), our filtering yields 212 clips (totaling \textbf{424} videos with reversals). This produces a low-ambiguity, psychophysically validated AoT evaluation benchmark that enables direct and reliable human–model comparison. Table~\ref{tab:motion_categories_figures} provides an overview of the motion categories with comparative statistics illustrating how \benchname~ filters the original dataset.

%This yields a low-ambiguity, psychophysically validated arrow-of-time evaluation benchmark that enables direct and reliable human–model comparison. The original dataset contains 720 videos (360 clips and their reversed counterparts). After filtering the cyclic motion category, \benchname comprises 212 clips, resulting in 424 videos when including both forward and reversed versions. Table~\ref{tab:motion_categories_figures} provides an overview of the motion categories with comparative statistics illustrating how \benchname~ filters the original dataset.

\begin{table*}[t]
\centering
\small
\setlength{\tabcolsep}{3pt}
\renewcommand\arraystretch{1.15}
\rowcolors{3}{black!3}{white}
\begin{tabularx}{\textwidth}{@{}
    >{\raggedright\arraybackslash}p{3.0cm}
    >{\centering\arraybackslash}p{1.8cm}
    >{\centering\arraybackslash}p{1.5cm}
    >{\centering\arraybackslash}p{1.4cm}
    >{\centering\arraybackslash}p{2.4cm}
    >{\centering\arraybackslash}p{2.2cm}
    >{\centering\arraybackslash}p{2.2cm}
    @{}}
\toprule
\textbf{Category \& Description} & \textbf{Reversal is easy for humans?} & \textbf{Human F1 (F/B)} & {\textbf{\# samples}} &
\textbf{Included in \benchname?} &
\textbf{Example 1} & \textbf{Example 2} \\
\midrule
\textbf{(1) Proceed:} forward locomotion of people, animals, or vehicles
& \yes & 86.5\,/\,82.5 & 82 & Yes
& \adjustbox{valign=m}{\includegraphics[width=2.2cm]{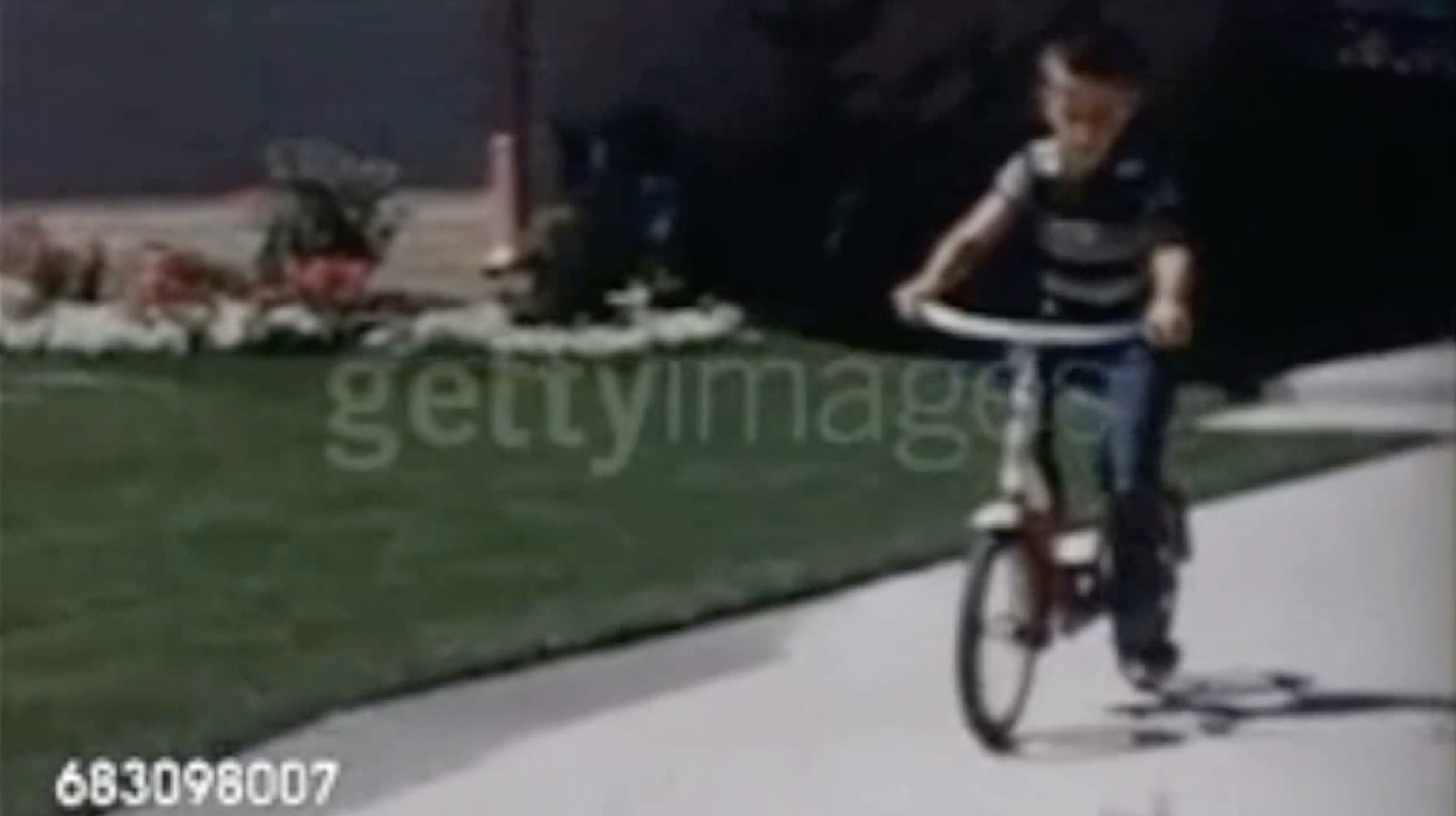}}
& \adjustbox{valign=m}{\includegraphics[width=2.2cm]{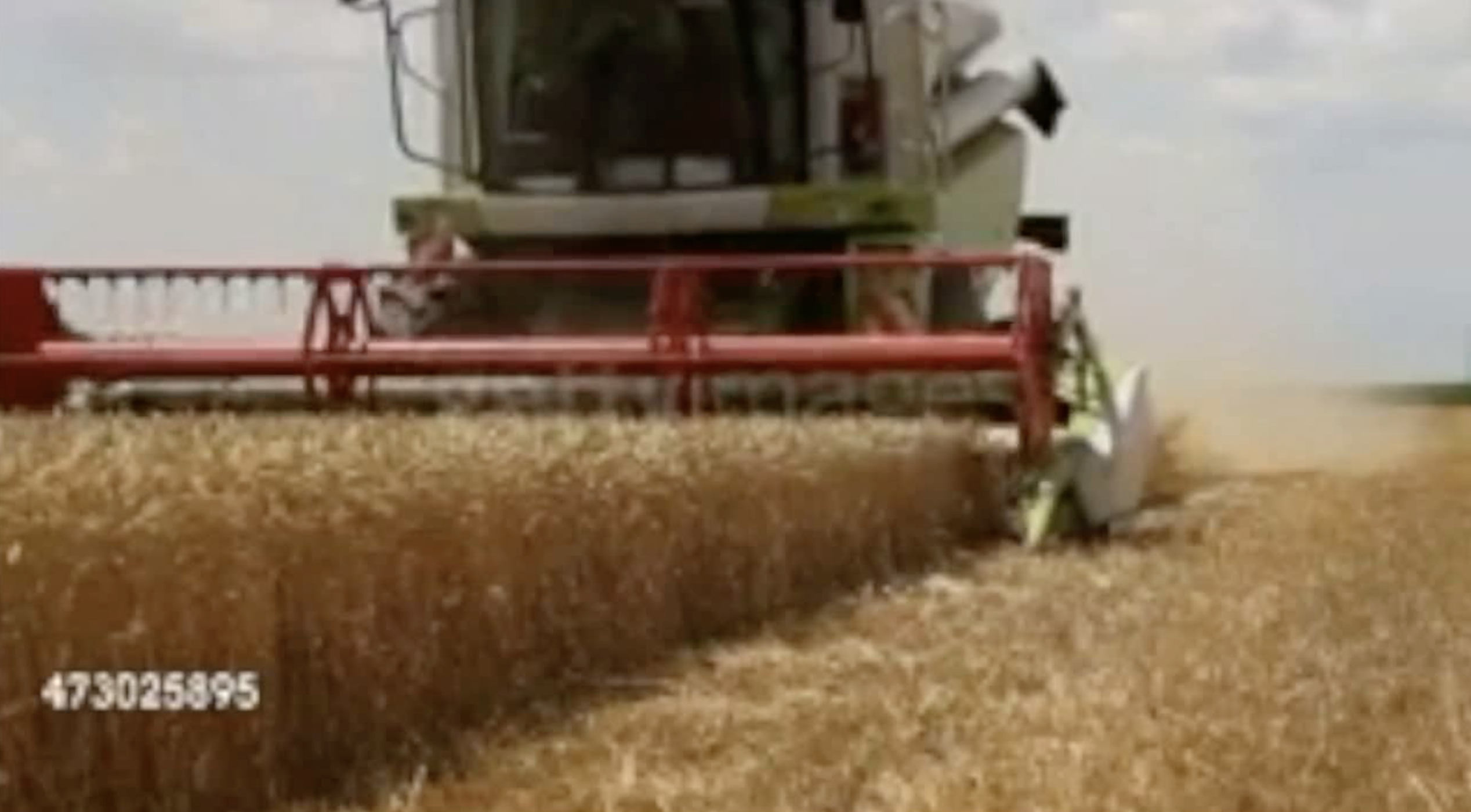}} \\
\textbf{(2) Fall:} free-fall/ballistic motion under gravity
& \yes & 86.9\,/\,82.8 & 84 & Yes
& \adjustbox{valign=m}{\includegraphics[width=2.2cm]{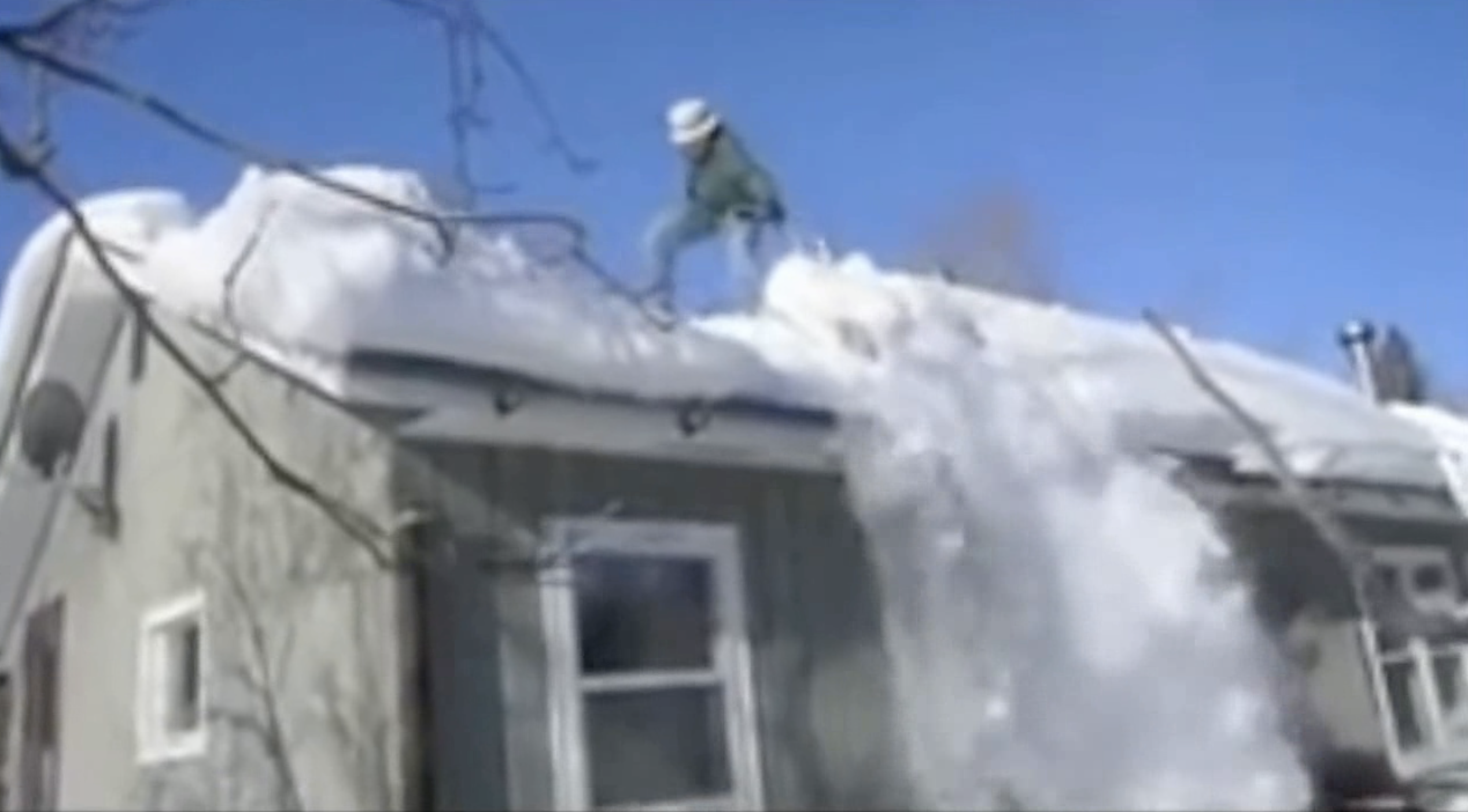}}
& \adjustbox{valign=m}{\includegraphics[width=2.2cm]{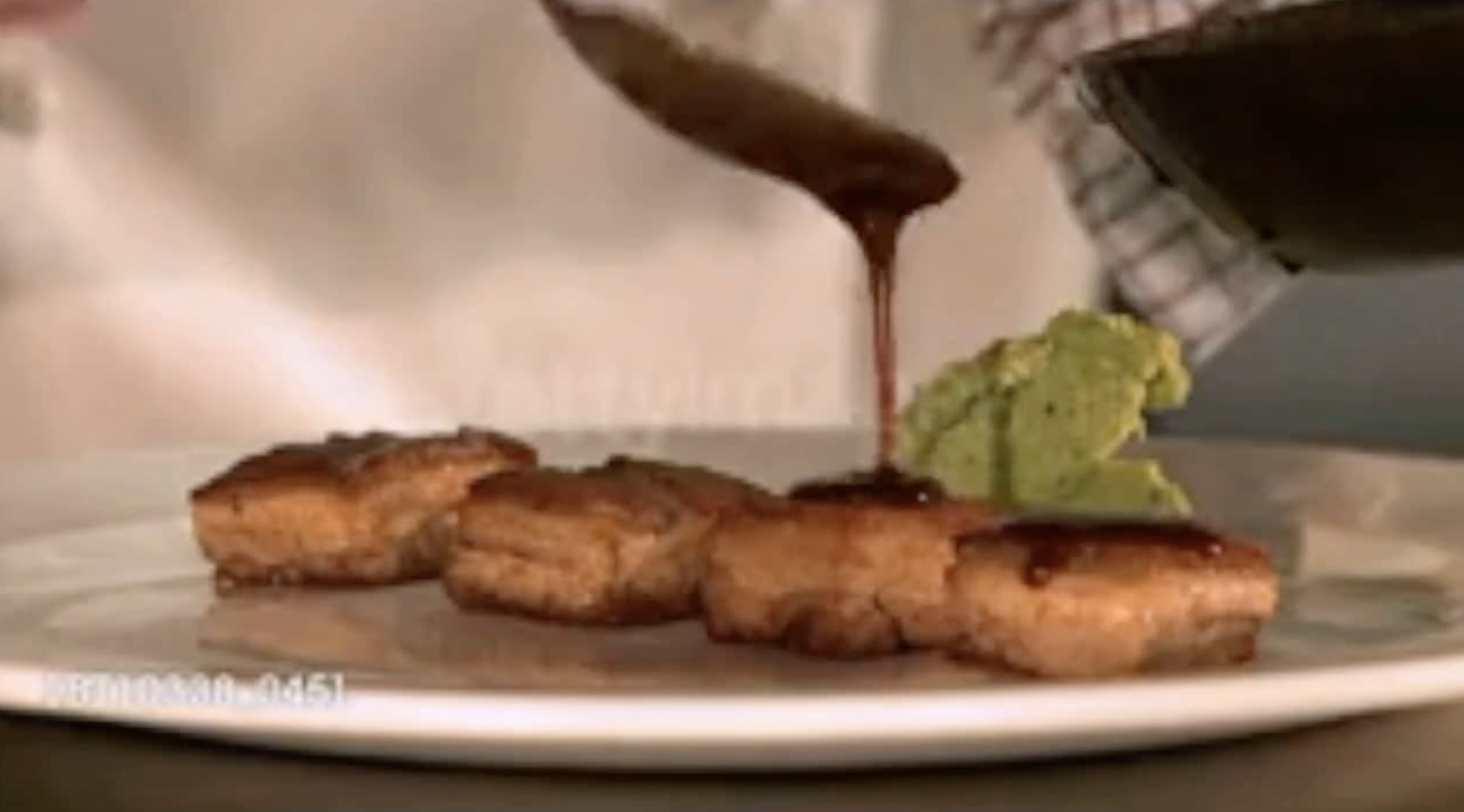}} \\
\textbf{(3) Diffusion:} centrifugal diffusion or small-particle explosions
& \yes & 84.6\,/\,78.7 & 56 & Yes
& \adjustbox{valign=m}{\includegraphics[width=2.2cm]{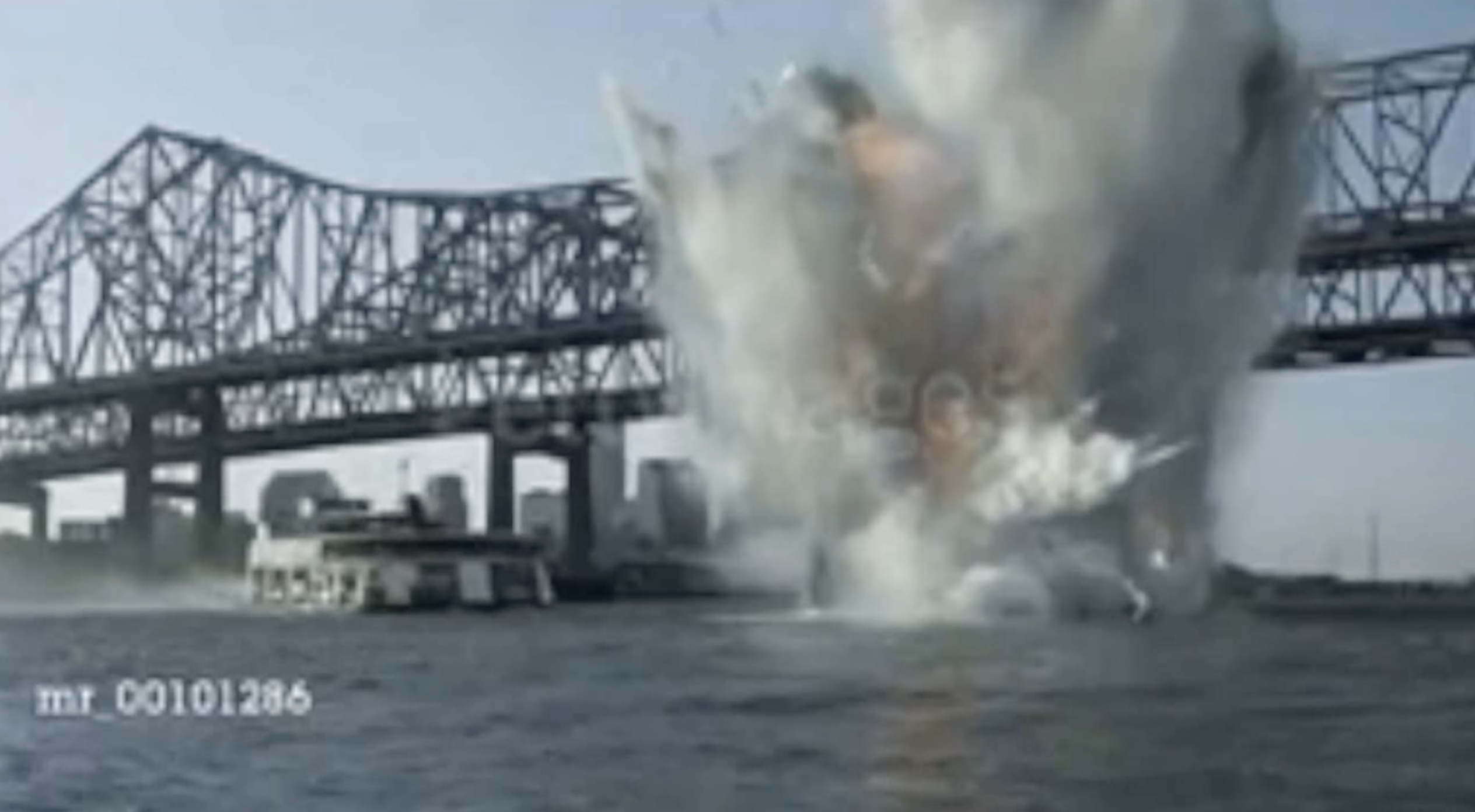}}
& \adjustbox{valign=m}{\includegraphics[width=2.2cm]{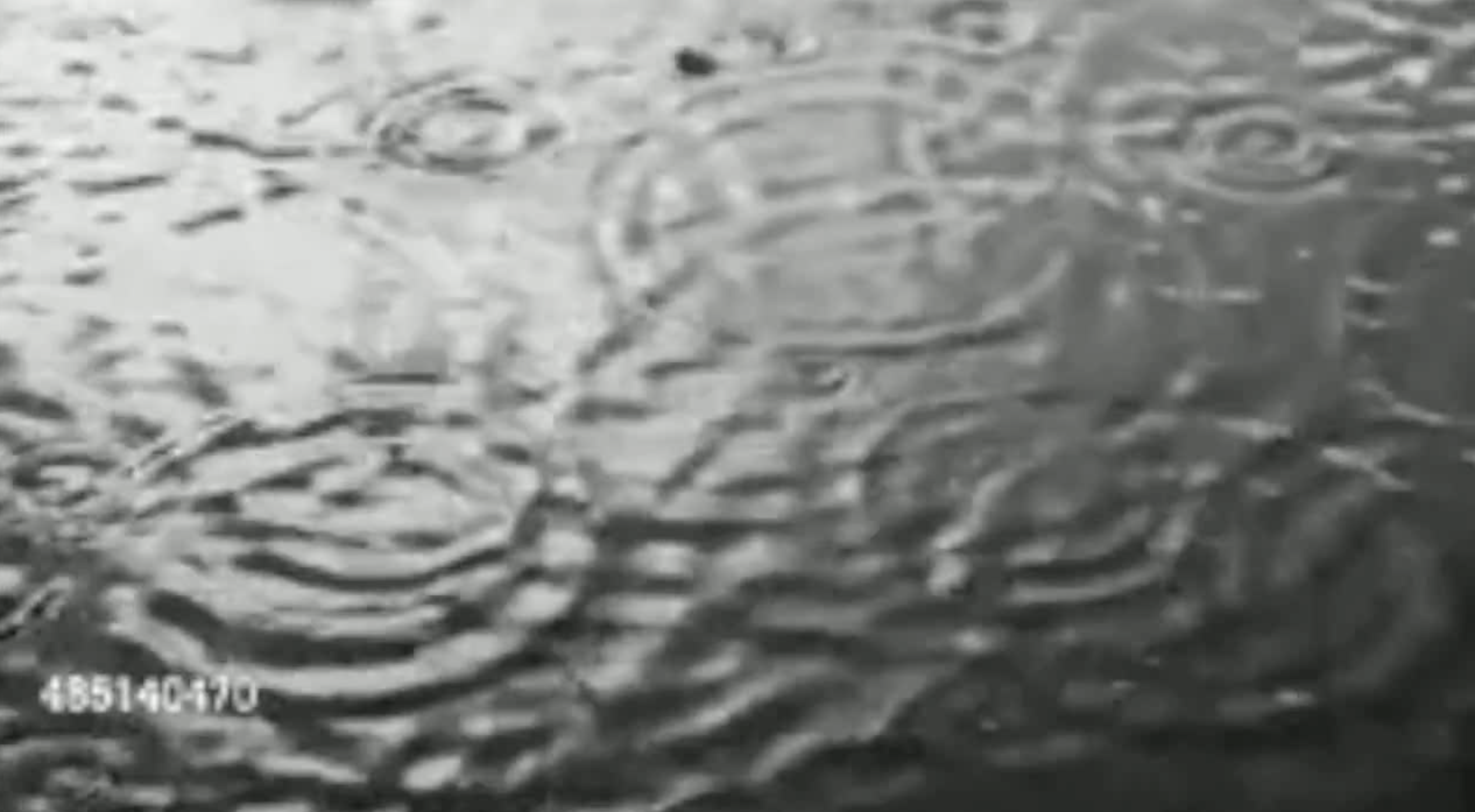}} \\
\textbf{(4) Division:} division of material by hand or tool
& \yes & 86.0\,/\,80.6 & 37 & Yes
& \adjustbox{valign=m}{\includegraphics[width=2.2cm]{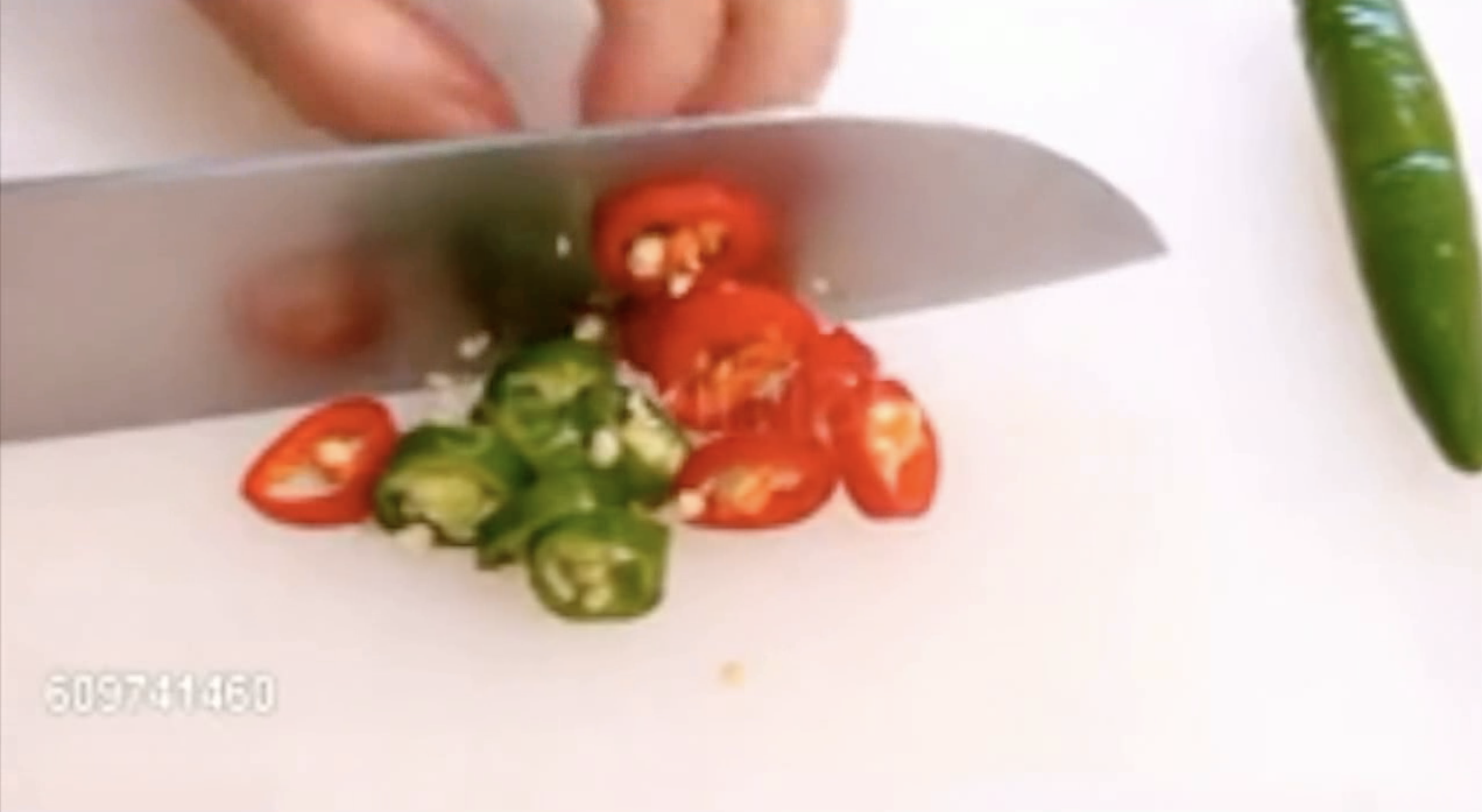}}
& \adjustbox{valign=m}{\includegraphics[width=2.2cm]{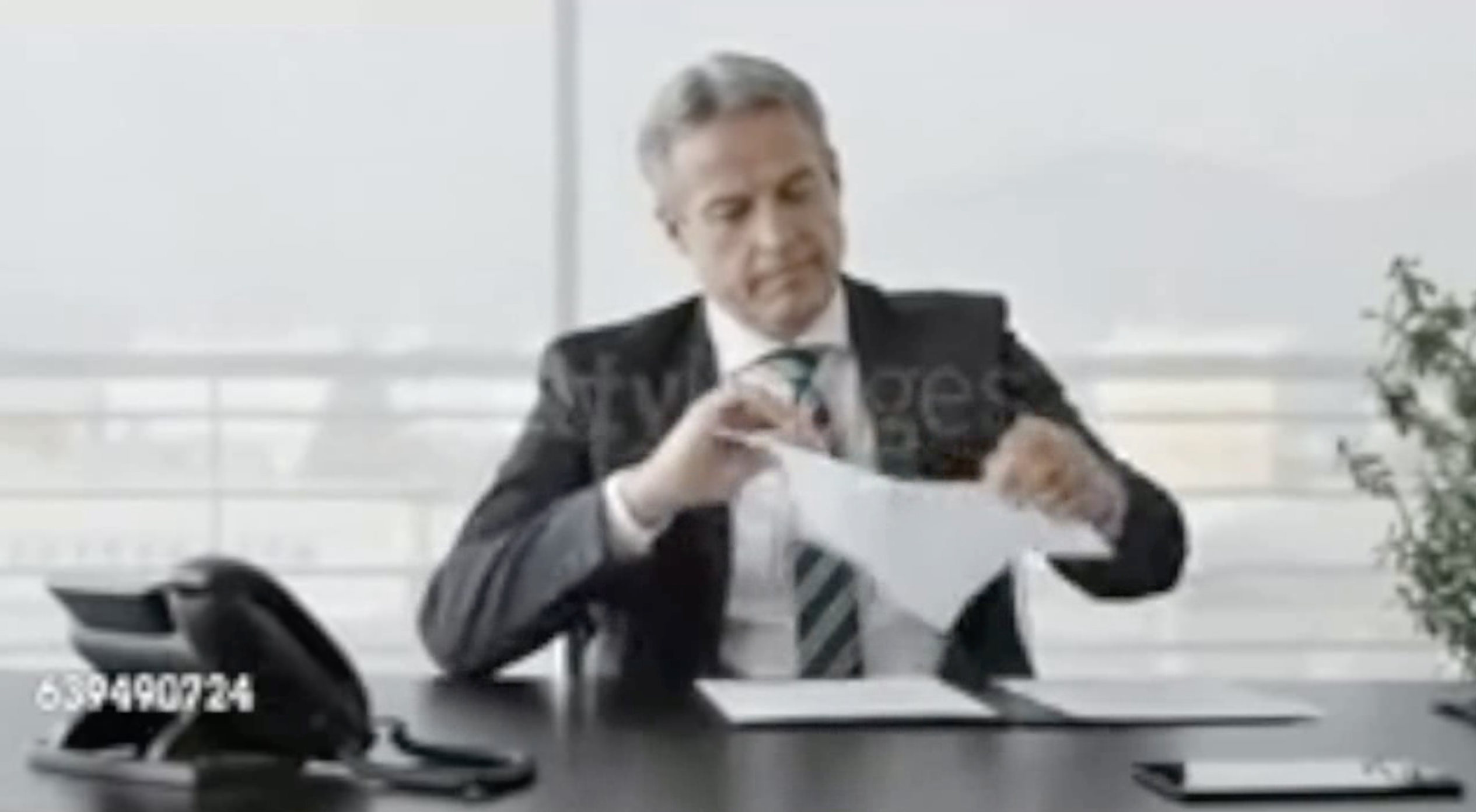}} \\
\textbf{(5) Put:} addition/construction of material by hand
& \yes & 84.1\,/\,77.4 & 67 & Yes
& \adjustbox{valign=m}{\includegraphics[width=2.2cm]{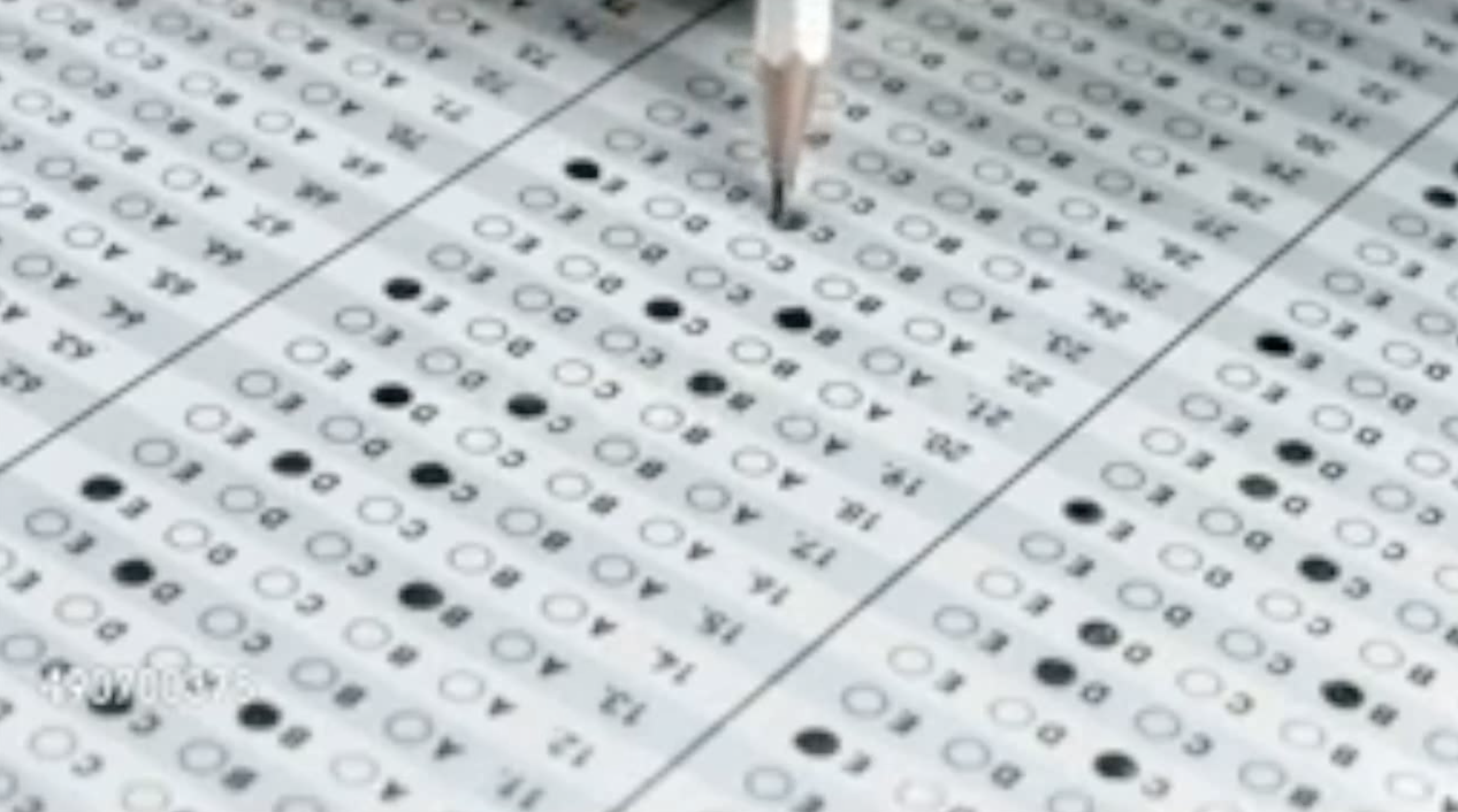}}
& \adjustbox{valign=m}{\includegraphics[width=2.2cm]{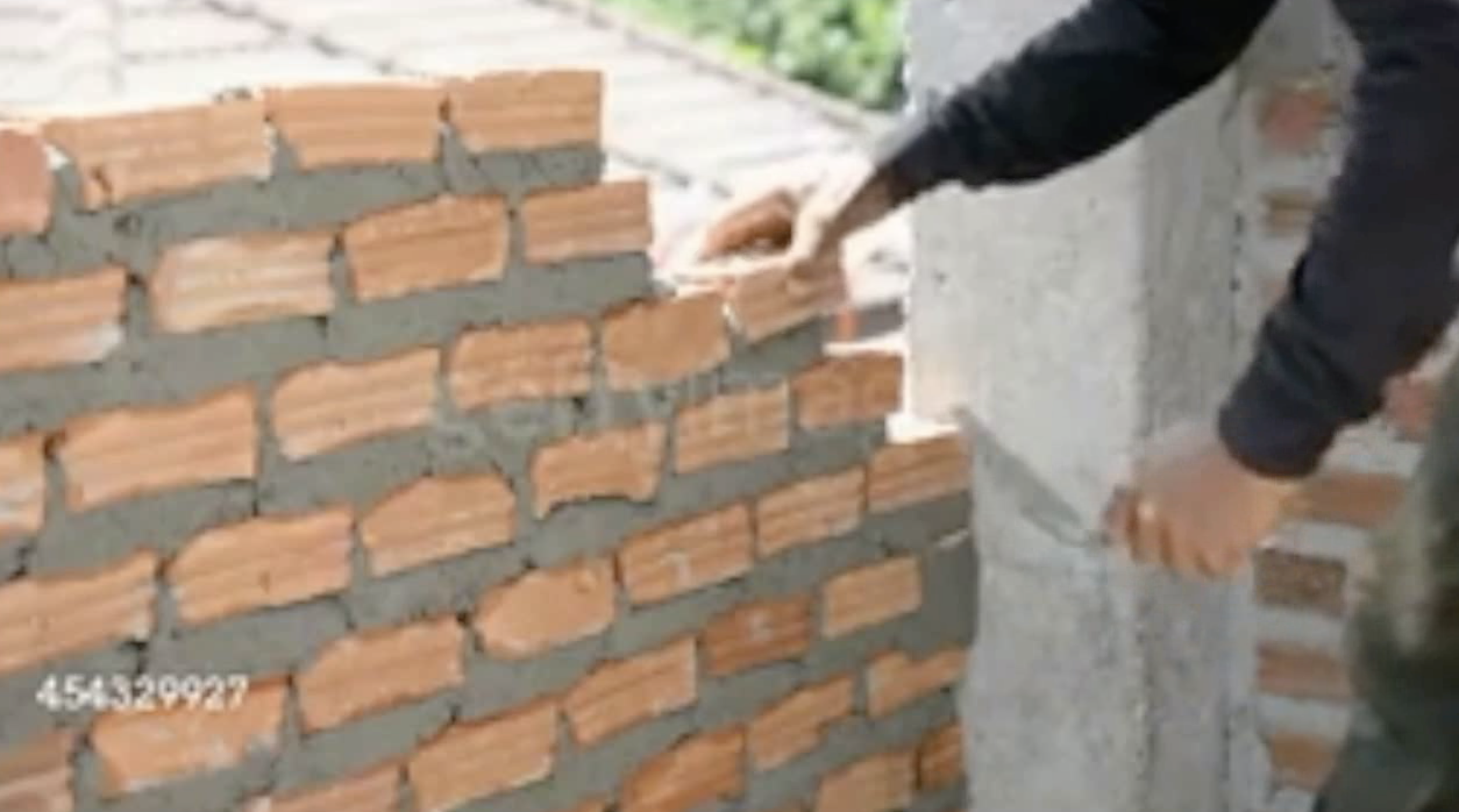}} \\
\textbf{(6) Reciprocal:} reciprocating (cyclic) motion
& \no & 71.6\,/\,\textbf{38.5} & 148 & \textbf{No}
& \adjustbox{valign=m}{\includegraphics[width=2.2cm]{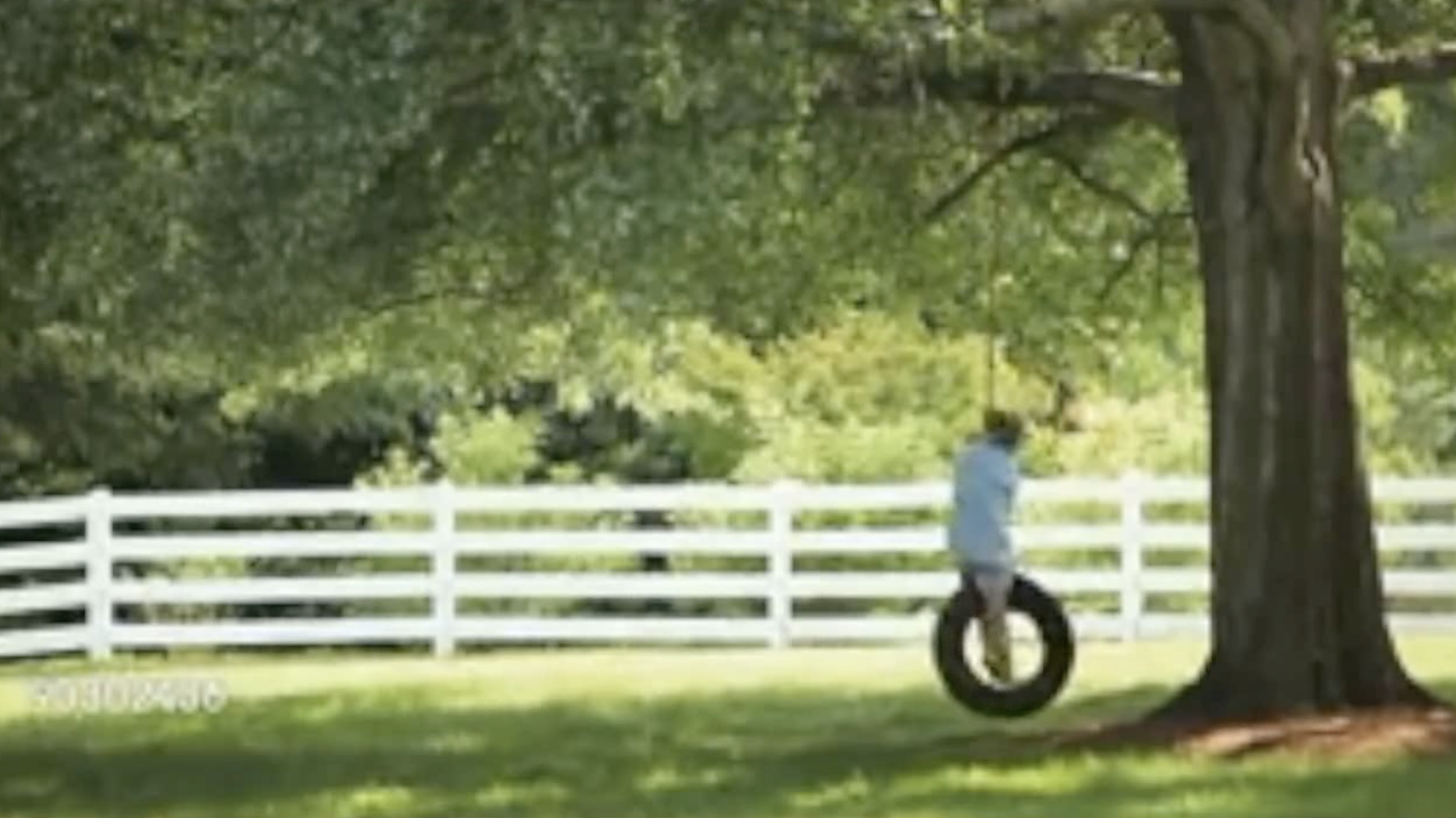}}
& \adjustbox{valign=m}{\includegraphics[width=2.2cm]{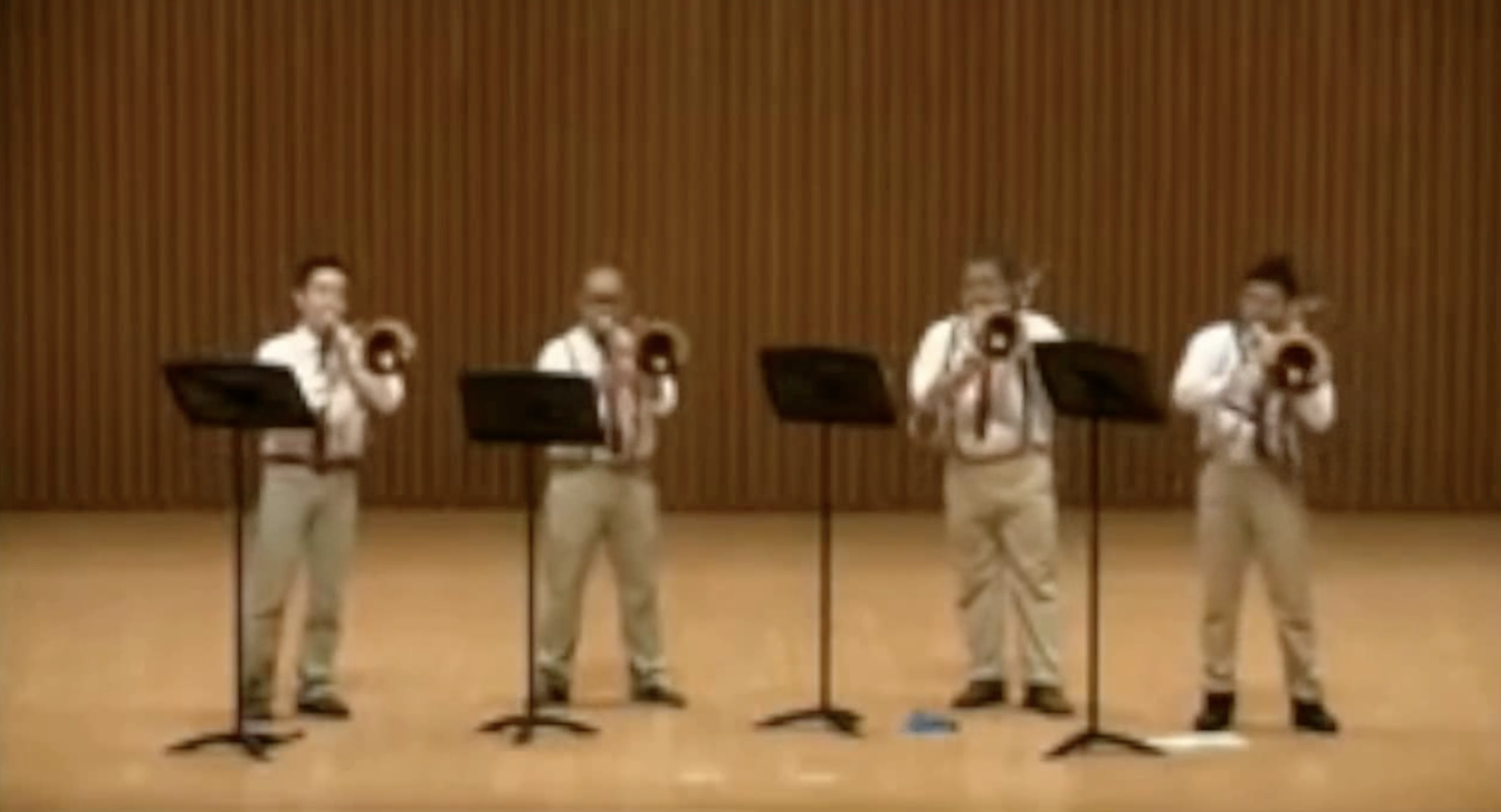}} \\
\bottomrule
\end{tabularx}
\caption{Motion categories following \citet{hanyu2023ready}, with human performance, counts, and visual examples.
Human F1 scores are reported as forward (F)/backward (B) for each motion category.
\benchname~ includes categories 1--5 (all clips) and excludes category 6.
\yes\ marks categories where reversal is typically easy for humans (irreversible processes); \no\ marks cyclic, bidirectional motions where reversal is challenging.}
\label{tab:motion_categories_figures}
\end{table*}

%\section{Task}
%We focus on the \emph{basic} AoT task: evaluating whether a VLM can discriminate forward from reversed playback. Specifically, given a video clip and a text prompt describing the task, the model outputs either \texttt{forward} or \texttt{backward}. The prompt instructs the model to determine the temporal direction of the video and may optionally include few-shot examples or chain-of-thought instructions, depending on the experimental setting, as detailed in Section~\ref{sec:exp}.

%\noindent\textbf{Evaluation.} We report accuracy overall, separately for forward versus reversed clips (to detect bias), and per motion category. Random guessing yields 50\%; human-level performance requires $\sim$75\% overall with category patterns matching human profiles.

\section{Experimental Settings}

\label{sec:exp}
%\subsection{Vision-Language Models in the Experiments}
We evaluate the following models (described in Section~\ref{subsec:vlms}) on \benchname: \textbf{Proprietary non-reasoning}: GPT-4o and GPT-4.1; \textbf{proprietary reasoning}: o3, o4-mini, GPT-5, and Gemini-2.5-Pro; \textbf{open-weight non-reasoning}: Qwen2-VL and Qwen2.5-VL; \textbf{open-weight reasoning}: Cosmos-Reason1-7B and QVQ-72B-Preview. We test these models across multiple experimental settings, which are detailed in the following sections.

\subsection{Zero-shot Settings} \label{sec:zero-shot-setting}

%In zero-shot experiments (Section~\ref{sec:zero-shot}), models receive video frames with a binary classification prompt requiring only \texttt{F} (forward) or \texttt{B} (backward) as output.

%\paragraph{System prompt.}
%\begin{quote}
%\textit{You will see videos played either forward or backward. Respond with only \texttt{F} (forward) or \texttt{B} (backward).}
%\end{quote}

%\paragraph{User prompt.}
%\begin{quote}
%\textit{[Video frames]}

%\textit{Determine whether this video plays forward or backward.}
%\end{quote}

%For zero-shot experiments (Section~\ref{sec:zero-shot}), we adopted a simple prompt design, asking the model to output only the \textit{Forward} or \textit{Backward} label. 

For the zero-shot experiments, we adopted a simple prompt design. In the system prompt, the model is asked to output only the \texttt{F} (forward) or \texttt{B} (backward) label as follows: 
%In the system prompt, we supplied the following instruction:
\vspace{4mm-\baselineskip}
\begin{quote}
\noindent\textbf{System prompt}
\textit{You will see videos provided from the user, played either forward or backward. 
Finish your answer with \texttt{F} or \texttt{B} only. 
\texttt{F} for forward and \texttt{B} for backward.}
\end{quote}
\vspace{4mm-\baselineskip}
%``
%You will see videos provided from the %user, played either forward or backward. 
%Finish your answer with \texttt{F} or %\texttt{B} only. 
%\texttt{F} for forward and \texttt{B} for backward.
%''

In the user prompt, we provide the sampled video frames with the following instruction:

%``
%Detect whether the video plays forward or backward with confidence.
%''
\vspace{4mm-\baselineskip}
\begin{quote}
\noindent\textbf{User prompt}
\textit{[Video frames]}
\textit{Detect whether the video plays forward or backward with confidence.}
\end{quote}
\vspace{0mm-\baselineskip}

\subsection{Few-shot Settings} \label{sec:few-shot-setting}
% For the non-chain-of-thought few-shot experiments (Section~\ref{sec:few-shot}), we reserved 4 videos as the shots: 2 forward and 2 backward videos from the benchmark, which contain visual cues that can be easily distinguished by human such as explosion. We excluded these 4 videos from the evaluation in this experiment. To craft the few-shots, we repeatedly append video and user prompt similarly as in Section~\ref{sec:zero-shot-setting} and a single label (''F`` or ''B``) to form a shot. 
%For the few-shot experiments (Section~\ref{sec:few-shot}), we selected four videos from the benchmark as few-shot examples: two forward and two backward videos. These videos were chosen because they contain visually distinctive cues such as explosions, which are easily recognizable by humans.

%For the few-shot experiments, we constructed demonstrations using four videos from the benchmark, two forward and two backward, each exhibiting visually distinctive temporal cues (such as explosions) for clear illustration. 
%To avoid data leakage, these four videos were excluded from the evaluation in experiments that require few-shot examples. Each few-shot example was constructed by sequentially appending the video and the corresponding user prompt (as described in Section~\ref{sec:zero-shot-setting}), followed by a single label in the assistant prompt (``F'' or ``B'') indicating the video direction.
For few-shot experiments, we constructed four demonstration examples from \benchname: two forward and two backward videos exhibiting visually distinctive temporal cues (e.g., explosions, falling objects). Each demonstration consists of sampled video frames, the user prompt from Section~\ref{sec:zero-shot-setting}, and the correct label (\texttt{F} for forward, \texttt{B} for backward). These four demonstration videos were excluded from the test set to prevent data leakage.

\subsection{Reasoning Effort Ablation Settings} \label{sec:cot_settings}

%In this setting, we provide the model with a sequence of examples, each of which shows
%how to solve the task step-by-step. The model then learns to generate its CoTs to solve it.
%Specifically, we reused two of the four videos employed in the non-CoT few-shot experiments (Section~\ref{sec:few-shot-setting}). We designed two variants of CoT prompts: Simple CoT and Multi-step CoT.
Reasoning-centric models are trained to generate internal reasoning chains automatically before producing outputs, while non-reasoning models usually respond directly unless explicitly prompted for step-by-step reasoning. 
%To control reasoning depth, we tuned the reasoning effort parameter for the former and manually designed CoT prompts for the latter.
We ablated the length of thinking of VLMs by either setting the reasoning effort parameter of reasoning models, or by manually prompting chain-of-thought (CoT) reasoning. 

\paragraph{Controlling Reasoning Effort for Proprietary Reasoning Models. } We set the reasoning effort in GPT-5 and Gemini-2.5-Pro to 3 levels: \textbf{low}, \textbf{medium}, and \textbf{high}. To be noted, the low, medium, and high reasoning effort of Gemini-2.5-Pro corresponds to the thinking budget in its native API of 1,024, 8,192, and 24,576 tokens, respectively. These values represent upper bounds; models do not necessarily use the full allocated budget.

\paragraph{Simulating Reasoning Effort Control with CoT Reasoning for Non-Reasoning Models. }
% In this setting, we provide models with demonstration examples that show step-by-step reasoning. The model then generates its own reasoning chain for test videos. 
%Open models do not offer an explicit way to control the reasoning effort. We therefore use few-shot CoT to induce similarly short or long reasoning contents.
%Specifically, we manually prompted an open non-reasoning model to reason step-by-step before predicting the final label. 
%We reused two of the four videos from few-shot experiments (Section~\ref{sec:few-shot-setting}) and designed two CoT variants: Simple CoT and Multi-Step CoT.

Open-weight models lack explicit reasoning effort control available in proprietary reasoning models. We therefore use few-shot chain-of-thought (CoT) prompting to simulate varying reasoning depths. Using two videos from Section~\ref{sec:few-shot-setting}, we created two CoT variants: \textbf{Simple CoT} with brief reasoning and \textbf{Multi-Step CoT} with elaborate step-by-step analysis. %This tests whether extended reasoning improves open models' performance analogously to reasoning effort parameters in proprietary systems.

\begin{figure}[!h]
    \centering
    \includegraphics[width=\columnwidth]{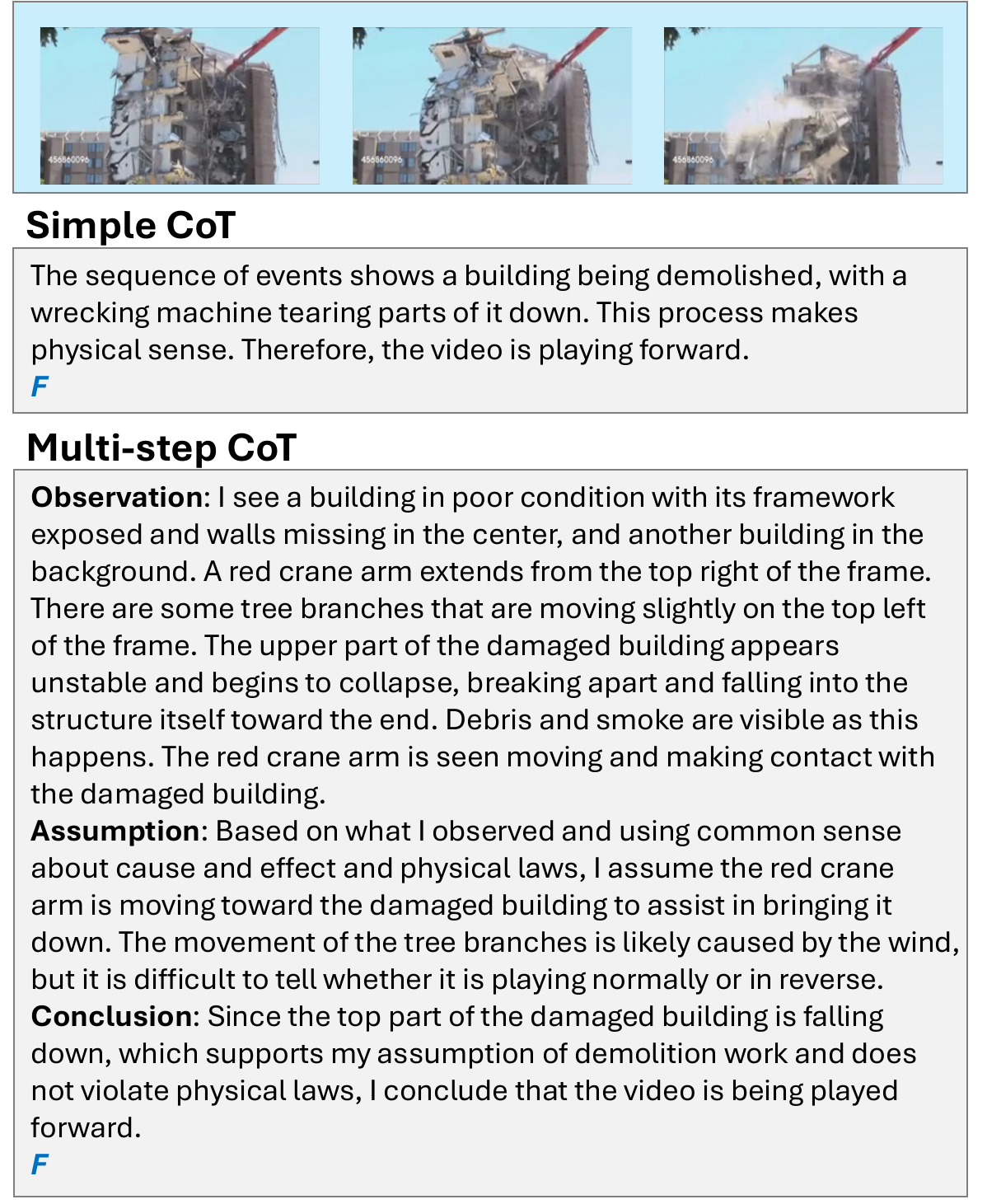}
    \caption{Manually curated chain-of-thought reasoning traces for a \textit{Forward} example. }
    \label{fig:forward_cots}
\end{figure}

\begin{figure}[!h]
    \centering
    \includegraphics[width=\columnwidth]{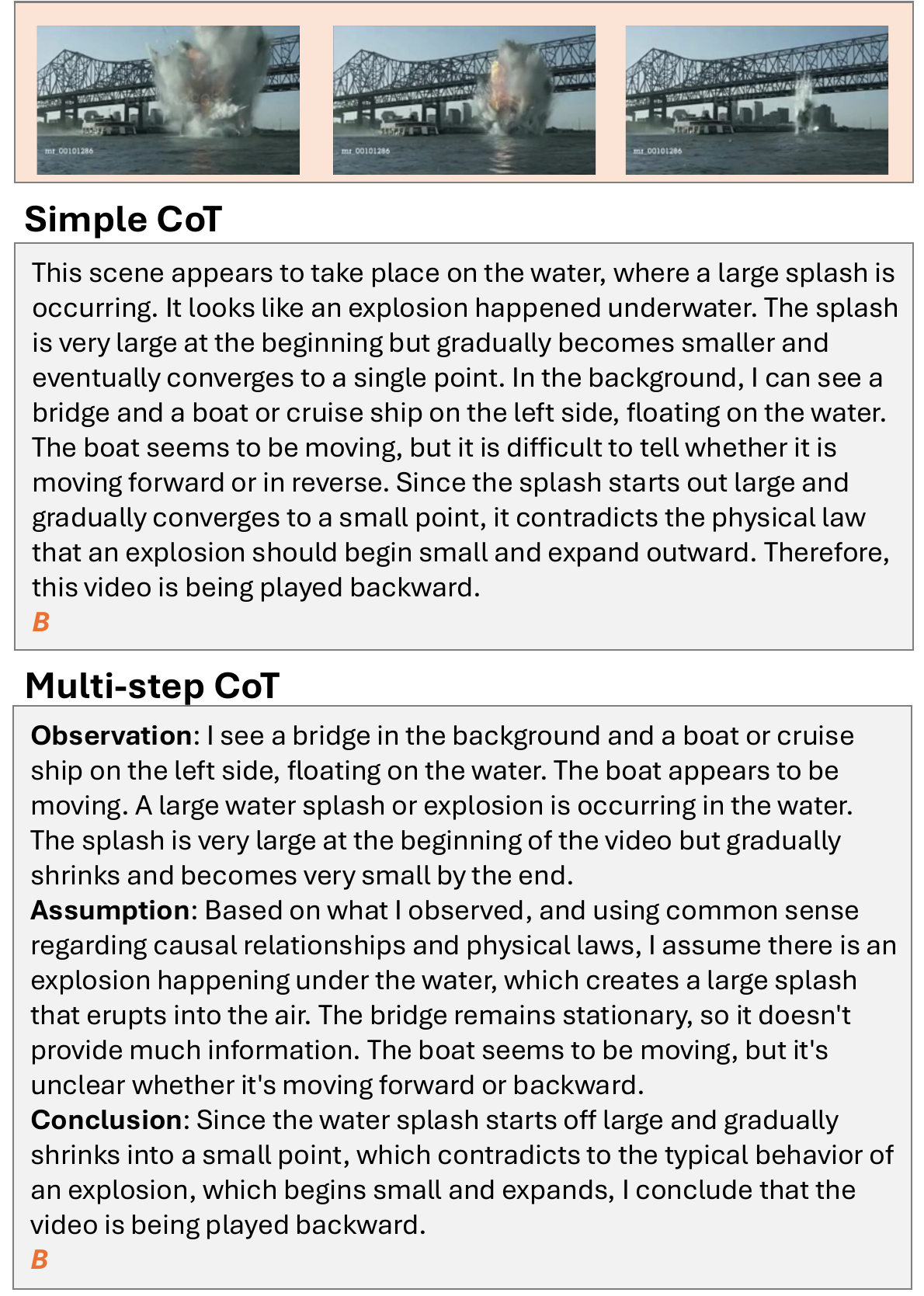}
    \caption{Manually curated chain-of-thought reasoning traces for a \textit{Backward} example. }
    \label{fig:backward_cots}
\end{figure}

\textbf{Simple CoT.} The model is instructed to focus on moving elements and identify visual cues indicating playback direction. We manually curated reasoning traces for each demonstration. The instruction for this setting is provided in Figure~\ref{fig:simple_cot_instruction}. Two manually curated reasoning traces are shown in Figure~\ref{fig:forward_cots},~\ref{fig:backward_cots}.
%An example is shown in Figure~\ref{fig:simple_cot}.

\textbf{Multi-step CoT.} We designed structured instructions with multi-step reasoning traces divided into three stages: \textit{Observation} (objectively describe visible events without assumptions), \textit{Assumption} (infer plausible physical or causal sequences), and \textit{Conclusion} (determine whether observations align with or contradict \textit{assumptions} to judge playback direction). The instruction for this setting is provided in Figure~\ref{fig:multi_step_cot_instruction}. Two manually curated reasoning traces are shown in Figure~\ref{fig:forward_cots},~\ref{fig:backward_cots}. A model output example is shown in Figure~\ref{fig:combined_armwrestling_reasoning}.
%\textbf{Simple CoT.} In this setting, the model is instructed to focus on the moving parts within the video and identify visual cues that may indicate the playback direction. We manually curated CoT traces as the textual reasoning output for each shot. An example of a Simple CoT prompt is shown in Figure~\ref{fig:simple_cot}.

%\textbf{Multi-step CoT.} In this setting, we designed structured instructions accompanied by multi-step CoT traces (Figure~\ref{fig:multi_step_cot_merged}). The reasoning process is divided into three stages:
%\textit{Observation}, where the model objectively describes visible events without making assumptions;
%\textit{Assumption}, where it infers plausible physical or causal sequences that would typically occur; and
%\textit{Conclusion}, where it determines whether the observed sequence aligns with or contradicts the assumed progression to decide if the video is played forward or backward.

\newlength{\ThumbH}

\subsection{Supervised Fine-Tuning  (SFT) Settings}

We constructed in-domain fine-tuning datasets to test whether explicit temporal direction supervision improves model performance. Note that all videos in \citet{hanyu2023ready}, \benchname, and these fine-tuning datasets are drawn from Moments in Time \cite{monfort2019moments}.  We randomly sampled 500, 1,500, and 2,500 clips distributed across its 339 action classes in the original dataset. These clips are entirely distinct from the videos in AoT-PsychBench. Each video was used to create two examples: the original (forward, labeled \texttt{F}) and its temporal reverse (backward, labeled \texttt{B}), yielding 1,000, 3,000, and 5,000 training examples.

We fine-tuned Qwen2-VL-7B and Qwen2.5-VL-7B using the zero-shot prompt format (Section~\ref{sec:zero-shot-setting}) with standard supervised learning. We then evaluated these models on \benchname.
 
\subsection{Evaluation Metrics}

We evaluate models using overall accuracy and F1 scores for Forward and Backward classes separately. Class-specific F1 scores highlight potential directional biases in model predictions.
%We evaluate the model's performance by calculating the F1 score for the Forward and Backward classes, respectively. In addition, we report the overall accuracy across all samples. %To gain deeper insights, we present detailed case studies in Section~\ref{sec:analysis}.

\subsection{Hyperparameter Settings}
\textbf{Frame Sampling and FPS Settings.}
Frames were sampled at equal temporal intervals without resizing. Frame rates were set based on model-specific defaults and training configurations: 2 FPS for Qwen2-VL, Qwen2.5-VL, and QVQ-72B-Preview (standard for these models), and 4 FPS for Cosmos-Reason1-7B (matching its inference settings recommendation\footnote{https://huggingface.co/nvidia/Cosmos-Reason1-7B}) and all proprietary models. Preliminary experiments showed that 4 FPS balances performance with API costs (Section~\ref{sec:fps_ablation_results}).

\textbf{Inference Parameters.}
For open-weight models (Qwen2-VL, Qwen2.5-VL, and QVQ-72B-Preview), we set temperature to 0.6 and top-p to 0.95. For Cosmos-Reason1-7B, we used temperature 0.6, top-p 0.95, and repetition penalty 1.05, consistent with settings reported in their paper. For GPT-4o (GPT-4o-2024-11-20) and GPT-4.1 (GPT-4.1-2025-04-14), we applied temperature 0.6, top-p 0.95, and frequency penalty 0. For OpenAI reasoning models (o3, o4-mini, GPT-5), inference parameters such as temperature and top-p are not exposed via the API and thus remained at default values. For Gemini-2.5-Pro (accessed 2025-10-02), we used the OpenAI-compatible API with default values: temperature 1.0 and top-p 0.95.

\textbf{SFT Parameters.} We froze the vision encoder and applied LoRA \cite{hu2021lora} (rank 8) to both the language model and the vision-language bridging MLP layer. We used a maximum sequence length of 2,048 tokens, a batch size of 2 per device with gradient accumulation over 2 steps (effective batch size of 4), a learning rate of $1 \times 10^{-5}$, and trained for 5 epochs with a cosine learning rate scheduler and 10\% warmup. %All training was performed in bfloat16 precision.

\section{Results}

% \input{LREC2026 Author's kit/tables/main_results}
% 20260314 matta: if input all at once, some will cluster at one page.
% moved some tables in 本文

\begin{table}[!tbp]
\centering
\begin{tabular}{@{}l c c c@{}}
\toprule
\textbf{Model} & \textbf{F. F1} & \textbf{B. F1} & \textbf{Acc.} \\
\midrule

\rowcolor{gray!15}\multicolumn{4}{c}{\textbf{Baselines}} \\
\addlinespace[2pt]
Random & — & — & 50.0 \\
% Human & 99.0 & 89.0 & 89.2 \\
Human & 90.0 & 88.1 & 89.2 \\

\rowcolor{gray!15}\multicolumn{4}{c}{\textbf{Open Models}} \\
\addlinespace[2pt]
\multicolumn{4}{l}{\textbf{Non-reasoning Models}} \\
Qwen2-VL-7B & 66.7 & 0.0 & 50.0 \\
Qwen2.5VL-7B & 63.0 & 19.5 & 49.3 \\
Qwen2.5VL-72B & 57.4 & 38.2 & 49.5 \\
% Qwen2-7B & 2 & 66.7 & 0.0 & 50.0 \\
% Qwen2.5VL-7B & 2 & 63.0 & 19.5 & 49.3 \\
% Qwen2.5VL-72B & 2 & 57.4 & 38.2 & 49.5 \\
\multicolumn{4}{l}{\textbf{Reasoning Models}} \\
\multirow{1}{*}{\shortstack[l]{QVQ-72B-Preview}}
  & 66.1 & 0.0 & 49.4 \\
 % & 2 & 66.1 & 0.0 & 49.4 \\
 % & 4 & 67.4 & 11.6 & 52.4 \\
\multirow{1}{*}{\shortstack[l]{cosmos-reason1 7B}} 
 % & 25.0 & 61.4 & 49.1 \\
 % & 2 & 25.0 & 61.4 & 49.1 \\
 & 31.2 & 63.3 & \textbf{52.1} \\
 % & 4 & 31.2 & 63.3 & 52.1 \\

\rowcolor{gray!15}\multicolumn{4}{c}{\textbf{Proprietary Models}} \\
\addlinespace[2pt]
\multicolumn{4}{l}{\textbf{Non-reasoning Models}} \\
GPT-4o & 65.4 & 24.9 & 52.6 \\
GPT-4.1 & 62.5 & 57.4 & \textbf{60.1} \\
\multicolumn{4}{l}{\textbf{Reasoning Models}} \\
o3 & 67.2 & 29.1 & 55.2 \\
o4-mini & 67.4 & 33.1 & 56.1 \\
GPT-5 & 68.7 & 26.8 & 56.1 \\
Gemini-2.5-pro & 65.9 & 51.4 & 59.9 \\
% GPT-4o & 4 & 65.4 & 25.9 & 52.6 \\
% GPT-4.1 & 4 & 60.3 & 49.3 & 55.5 \\
% o3 & 4 & 67.2 & 29.1 & 55.2 \\
% o4-mini & 4 & 67.4 & 33.1 & 56.1 \\
% GPT-5 & 4 & 68.7 & 26.8 & 56.1 \\
% Gemini-2.5-pro & 4 & 65.9 & 51.4 & \textbf{59.9} \\
\bottomrule
\end{tabular}
\caption{Zero-shot performance.}
\label{tab:zero_shot_results}
\end{table}

\subsection{Zero-shot Performance}\label{sec:zero-shot}

Table~\ref{tab:zero_shot_results} presents zero-shot results, revealing two key gaps: \textbf{open vs. proprietary} and \textbf{VLM vs. human} performance gaps. Additionally, we did not find advantages of reasoning models over non-reasoning models.

\paragraph{Open vs. Proprietary. }
\textbf{Open-weight models} cluster around 50\% accuracy (random baseline), with Cosmos-Reason1-7B slightly higher at 52.1\%. QVQ-72B-Preview, designed for mathematical reasoning over single images rather than physical dynamics, performs at chance with extreme label bias (Section~\ref{sec:label_prediction_bias}). Cosmos-Reason1-7B, despite being trained on an undisclosed AoT dataset where it reportedly achieved 60\% accuracy, shows minimal improvement on our benchmark.
\textbf{Proprietary models} consistently exceed random guessing, with GPT-4.1 achieving the highest zero-shot performance (60.1\%). 

\paragraph{VLM vs. Human. }
All VLMs lag substantially behind humans (89.2\%), with a gap of 29.1 percentage points even for the best zero-shot model, GPT-4.1.

\paragraph{Reasoning vs. Non-Reasoning:} Explicit reasoning capability provides no clear advantage. The non-reasoning model GPT-4.1 (60.1\%) outperforms reasoning models o3 (55.2\%), o4-mini (56.1\%), and GPT-5 (56.1\%). %and benefits more from few-shot prompting (Section~\ref{sec:few-shot}). 

\begin{table}[!tbp]
\centering
\begin{tabular}{@{}l c c c c@{}}
\toprule
\textbf{Model} & \textbf{\#Shots} & \textbf{F. F1} & \textbf{B. F1} & \textbf{Acc.} \\
\midrule
% \rowcolor{gray!15}\multicolumn{6}{c}{\textbf{Baselines}} \\
% \addlinespace[2pt]
% Random & — & — & — & — & 50.0 \\
% Human & — & 30 & 99.0 & 89.0 & 89.2 \\
\rowcolor{gray!15}\multicolumn{5}{c}{\textbf{Open Model}} \\
\addlinespace[2pt]
\multirow{3}{*}{\shortstack[l]{Qwen2.5\\VL-72B}} 
% eval on 420
 & 0 & 57.4 & 38.0 & 49.5 \\
 % & 2 & 64.3 & 19.2 & 50.5?? \\
 & 2 & 64.2 & 18.7 & 50.2 \\
 & 4 & 65.4 & 21.1 & 51.9 \\
 % & 0 & 2 & 64.2 & 18.5 & 50.2 \\ % these are CoT results
 % & 2 & 2 & 67.3 & 3.76 & 51.2 \\
 % & 2 & 2 & 66.7 & 5.43 & 50.7 \\

\rowcolor{gray!15}\multicolumn{5}{c}{\textbf{Proprietary Models}} \\
\addlinespace[2pt]
\multirow{3}{*}{GPT-4.1} 
 % & 0 & 4 & 57.3 & 47.9 & 53.1 \\ # full neuro paper data
 % w/ 4 demonstrations
 % & 0 & 4 & 60.5 & 49.5 & 55.7 \\
 % & 2 & 2 & 60.6 & 51.2 & 56.4 \\
 % & 4 & 2 & 61.5 & 50.9 & 56.8 \\
 % & 2 & 4 & 63.6 & 57.7 & \textbf{60.9} \\
 % & 4 & 4 & 61.1 & 52.7 & 57.3 \\
 % & 2 & 8 & 60.7 & 52.0 & 56.7 \\

 % w/o 4 demonstrations
 & 0 & 62.3 & 57.4 & \textbf{60.0} \\
 % & 2 & 2 & 60.6 & 51.2 & 56.4 \\
 % & 4 & 2 & 61.3 & 50.8 & 56.7 \\
 & 2 & 59.5 & 51.7 & 56.0 \\
 & 4 & 60.0 & 49.9 & 55.5 \\
 % & 2 & 8 & 60.6 & 51.3 & 56.5 \\
\midrule
\multirow{3}{*}{GPT-4o}
 % w/ 4 demonstrations
 % & 0 & 4 & 65.5 & 25.4 & 52.8 \\
 % & 2 & 4 & 66.9 & 5.45 & 50.9 \\
 % & 4 & 4 & 66.7 & 8.81 & 51.2 \\

 % w/o 4 demonstrations
 & 0 & 65.4 & 24.9 & 52.6 \\
 & 2 & 66.8 & 4.61 & 50.7 \\
 & 4 & 66.5 & 5.07 & 50.7 \\
\midrule

 \multirow{3}{*}{GPT-5} 
 % w/o 4 demonstrations
 & 0 & 68.7 & 26.8 & 56.1 \\ % all with medium reasoning
 % & 2 & 2 & 67.9 & 29.7 & 56.0 \\
 % & 4 & 2 & 68.2 & 20.9 & 56.4 \\
 & 2 & 69.5 & 29.2 & 57.4 \\
 & 4 & 69.2 & 30.9 & 57.4 \\

\bottomrule
\end{tabular}
\caption{Few-shot performance. }
\label{tab:few-shot-results}
\end{table}

\subsection{Few-shot Performance} \label{sec:few-shot}
We test whether few-shot examples elicit in-context learning for AoT (Table~\ref{tab:few-shot-results}). Overall, few-shot prompting does \emph{not} consistently improve performance and often amplifies label prediction bias. GPT-5 shows modest, stable gains (Accuracy: 56.1→57.4/57.4\%; Forward F1: 68.7→69.5/69.2; Backward F1: 26.8→29.2/30.9). In contrast, GPT-4.1 degrades with few-shot prompts (Accuracy: 60.0→56.0/55.5\%; Forward F1: 62.3→59.5/60.0; Backward F1: 57.4→51.7/49.9). GPT-4o shows severe degradation: accuracy drops (52.6→50.7\%), with Backward F1 collapsing catastrophically (24.9→4.6/5.1), indicating near-complete failure to detect reversed videos. The open model Qwen2.5-VL-72B exhibits minimal accuracy change (49.5→51.9\%) but substantially worsened bias (Backward F1: 38.0→18.7/21.1). These results demonstrate that few-shot prompting provides limited benefits for temporal reasoning: only one model (GPT-5) shows modest improvement, while others either decline or amplify their existing forward bias, suggesting that in-context learning is insufficient to elicit robust AoT understanding.
%%We test whether few-shot examples elicit in-context learning in VLMs and improve AoT performance (Table~\ref{tab:few-shot-results}). We find: \textbf{(1)} Few-shot prompting is effective for GPT-4.1. Its overall accuracy reaches 60.5\% (+5.0), and its label prediction bias is modestly reduced (+7.9 \textit{Backward} F1 with 2 shots). GPT-5 shows a similar pattern: slight accuracy gains and reduced bias (+1.3 accuracy, +4.1 \textit{Backward} F1  with 4 shots). This suggests that models already performing above random on AoT can benefit from few-shot prompts. \textbf{(2)} In contrast, Qwen2.5-VL-72B and GPT-4o start from weaker baselines, and using 2 or 4 shots did not bring significant improvement on accuracy while amplifying label prediction bias.
% We examine whether few-shot examples enable the in-context learning ability of VLMs and improve performance on AoT, shown in Table~\ref{tab:few-shot-results}. We observed that (1) few-shot is very effective on GPT-4.1. Not only does it achieve a 60.5 (+5.0) overall accuracy, but its label prediction bias also mitigated a little. Similar results can be seen on GPT-5, too: few-shot slightly improved its performance and helped its label prediction bias. This suggests that models that can already perform above-random on AoT can benefit from few-shot prompting. (2) On the other hand, Qwen2.5VL-72B and GPT-4o initially had subpar performance, and after incorporating 2 shots or 4 shots, their performance even degraded, and their label prediction bias significantly magnified. 

\begin{table}[!tbp]
\centering
\begin{tabular}{@{}l c c c c@{}}
\toprule
\textbf{Model} & \textbf{Effort} & \textbf{F. F1} & \textbf{B. F1} & \textbf{Acc.} \\
\midrule
% \rowcolor{gray!15}\multicolumn{6}{c}{\textbf{Baselines}} \\
% \addlinespace[2pt]
% Random & — & — & — & — & 50.0 \\
% Human & — & 30 & 99.0 & 89.0 & 89.2 \\

% \rowcolor{gray!15}\multicolumn{6}{c}{\textbf{Open Models}} \\
% \addlinespace[2pt]
% \multirow{2}{*}{\shortstack[l]{QVQ-72B\\Preview}} 
%   & — & 2 & 66.1 & 0.0 & 49.4 \\
%   & — & 4 & 67.4 & 11.6 & 52.4 \\
% \midrule
% \multirow{2}{*}{\shortstack[l]{cosmos-\\reason1 7B}} 
%   & — & 2 & 25.0 & 61.4 & 49.1 \\
%   & — & 4 & 31.2 & 63.3 & 52.1 \\

\rowcolor{gray!15}\multicolumn{5}{c}{\textbf{Proprietary Models}} \\
\addlinespace[2pt]
% o3 & III & 67.2 & 29.1 & 55.2 \\
% \midrule
% o4-mini & III & 67.4 & 33.1 & 56.1 \\
% \midrule
\multirow{3}{*}{GPT-5}
  % & I & 66.5 & 7.11 & 50.7 \\
  & Low & 68.9 & 34.4 & \textbf{57.8} \\
  & Medium & 68.7 & 26.8 & 56.1 \\
  & High & 69.1 & 25.7 & 56.4 \\
\midrule
\multirow{3}{*}{\shortstack[l]{Gemini-\\2.5-pro}}
  & Low & 67.2 & 52.1 & \textbf{61.1} \\
  & Medium & 65.9 & 51.4 & 59.9 \\
  & High & 65.3 & 49.0 & 58.7 \\

\bottomrule
\end{tabular}
\caption{Reasoning effort ablation. }
\label{tab:reasoning_results}
\end{table}

\begin{table}[!tbp]
\centering 
\begin{tabular}{@{}l c c c c@{}}
\toprule
\textbf{Setting} & \textbf{Shots} & \textbf{F. F1} & \textbf{B. F1} & \textbf{Acc.} \\
\midrule

% \rowcolor{gray!15}\multicolumn{6}{c}{\textbf{Baselines}} \\
% \addlinespace[2pt]
% Random & — & — & — & — & 50.0 \\
% Human & — & 30 & 99.0 & 89.0 & 89.2 \\

\rowcolor{gray!15}\multicolumn{5}{c}{\textbf{Qwen2.5VL-72B-instruct}} \\
\addlinespace[2pt]
\multirow{1}{*}{No CoT} 
% eval on 420
  % & 0 & 57.4 & 38.0 & 49.5 \\
  & 2 & 64.2 & 18.7 & 50.2 \\
\midrule
\multirow{1}{*}{Simple} 
  % & 0 & 64.2 & 18.5 & 50.2 \\
  & 2 & 67.3 & 3.76 & 51.2 \\
\midrule
 % eval on 420
Multi-step & 2 & 66.7 & 5.48 & 50.7 \\

\bottomrule
\end{tabular}
\caption{Chain-of-Thought performance on the Qwen2.5VL-72B-instruct model. }

\vspace{-3mm}

\label{tab:cot_results}
\end{table}

\begin{figure*}[!th]
%================== layout controls ==================%
% \newlength{\LeftW}       
\setlength{\LeftW}{0.20\linewidth} % left column width
% \newlength{\GapW}        
\setlength{\GapW}{0.75em}           % gap between columns

% Boxes for the two right-side text panels
\newsavebox{\PromptBoxA}
\newsavebox{\PromptBoxB}

%================== 1) typeset the two listings into boxes ==================%
\begin{lrbox}{\PromptBoxA}
  \begin{minipage}[t]{\dimexpr\linewidth-\LeftW-\GapW\relax}
\begin{lstlisting}[style=promptbox]
<<bf>>An incorrect output from Qwen2.5VL-72B with Multi-step CoT prompting:<</bf>>
<<bf>>Observation<</bf>>: The video shows <<green>>two men engaged in arm-wrestling at a desk<</green>>. (......) <<orange>>Their hands are clasped together on the desk, and they are pulling against each other. <</orange>>
<<bf>>Assumption<</bf>>: In a typical arm-wrestling match, <<green>>the participants start with their hands clasped and then exert force to try to pin the opponent's arm to the table<</green>>. (......)
<<bf>>Conclusion<</bf>>: The video shows the men in the middle of the struggle, with their hands clasped and both leaning forward in an effort to win. <<red>>This matches the expected sequence of a forward-playing arm-wrestling match. There are no signs of reversed motion, such as the hands separating or the men leaning backward unexpectedly<</red>>. Therefore, the video is played <<red>>forward<</red>>.
\end{lstlisting}
  \end{minipage}
\end{lrbox}

\begin{lrbox}{\PromptBoxB}
  \begin{minipage}[t]{\dimexpr\linewidth-\LeftW-\GapW\relax}
\begin{lstlisting}[style=promptbox]
<<bf>>Gemini-2.5-Pro Self-Explained Reasoning:<</bf>>
The video shows two men engaged in <<green>>an arm wrestling match<</green>>.
1. <<bf>>Normal Progression of Arm Wrestling<</bf>>: <<green>>In a standard arm wrestling match, the contest begins with both participants' arms in an upright, near-vertical position<</green>>. The goal is to force the opponent's arm down onto the table. (......)
2. <<bf>>Observed Motion in the Video<</bf>>: The video clip <<green>>starts with the man in the white shirt in a clear losing position, with his arm almost completely pinned to the desk<</green>>. As the video plays, his arm moves upward and away from the desk, (......)
3. <<bf>>Conclusion<</bf>>: <<green>>The action shown is the exact reverse of how an arm wrestling match progresses<</green>>. A match ends with an arm going down, it doesn't start there and move up. Because the video shows the conclusion of the match first and then moves back toward the starting position, <<green>>it is playing in reverse<</green>>.
\end{lstlisting}
  \end{minipage}
\end{lrbox}

%================== 2) measure heights and set total ==================%
\newlength{\PromptHA} \setlength{\PromptHA}{\dimexpr\ht\PromptBoxA+\dp\PromptBoxA\relax}
\newlength{\PromptHB} \setlength{\PromptHB}{\dimexpr\ht\PromptBoxB+\dp\PromptBoxB\relax}
\newlength{\PromptTotal} \setlength{\PromptTotal}{\dimexpr\PromptHA+\PromptHB+\GapW\relax}

% Thumbnail height on the left (3 frames stacked)
\setlength{\ThumbH}{0.33\PromptTotal}

%================== 3) place left images + right stacked boxes ==================%
\centering
\noindent
\begin{minipage}[t][\PromptTotal][t]{\LeftW}   % enforce same total height
  \vspace{0pt} % ensure consistent top alignment
  \setlength{\parskip}{0pt}%
  \includegraphics[height=\ThumbH,width=\linewidth,keepaspectratio]{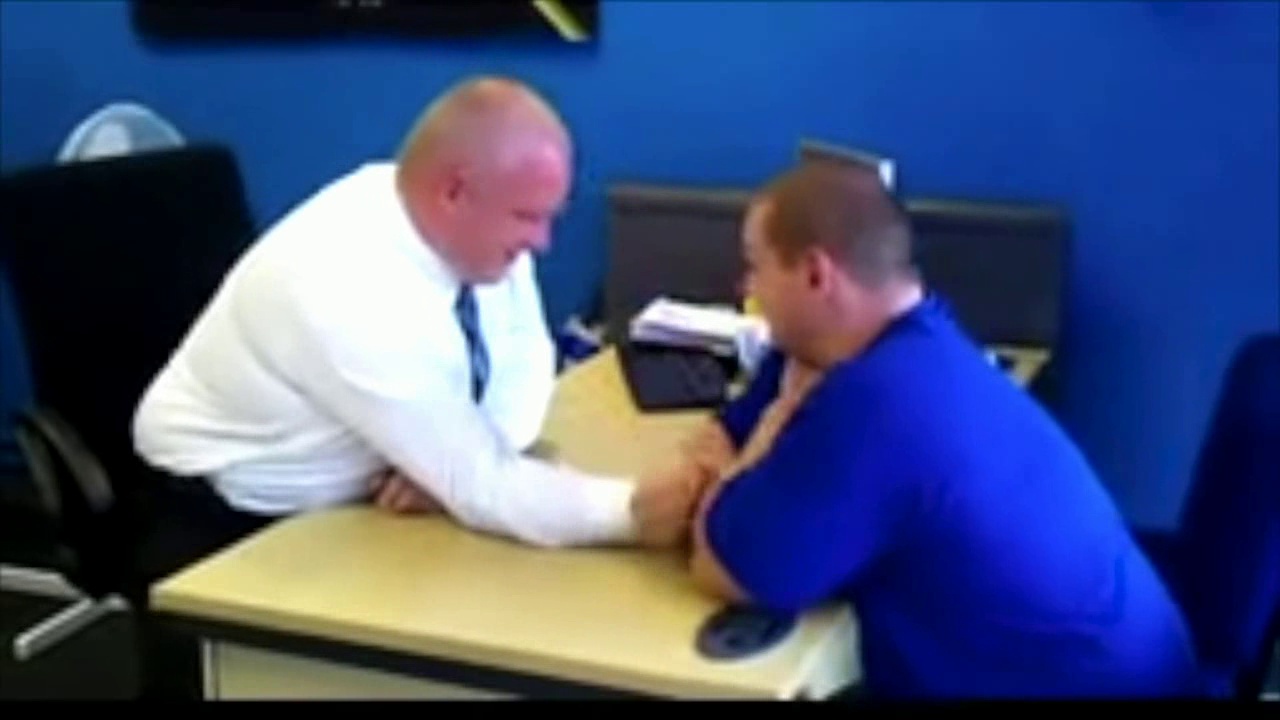}\par
  \includegraphics[height=\ThumbH,width=\linewidth,keepaspectratio]{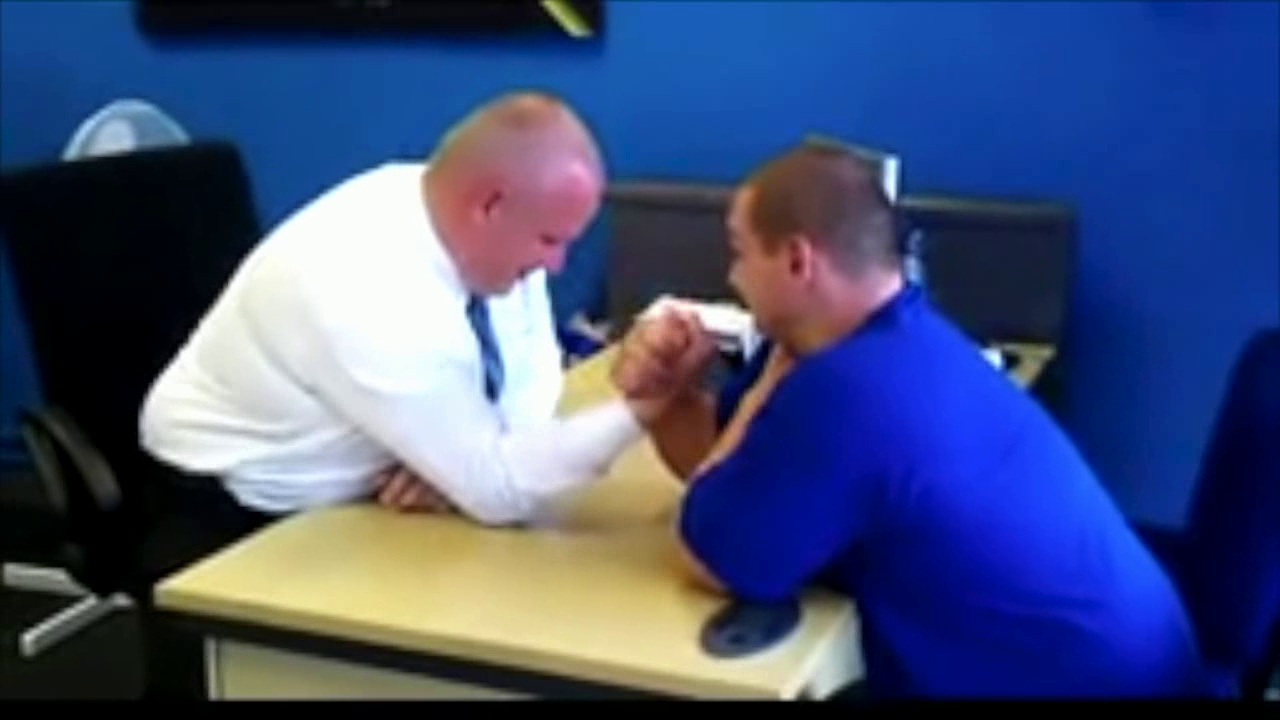}\par
  \includegraphics[height=\ThumbH,width=\linewidth,keepaspectratio]{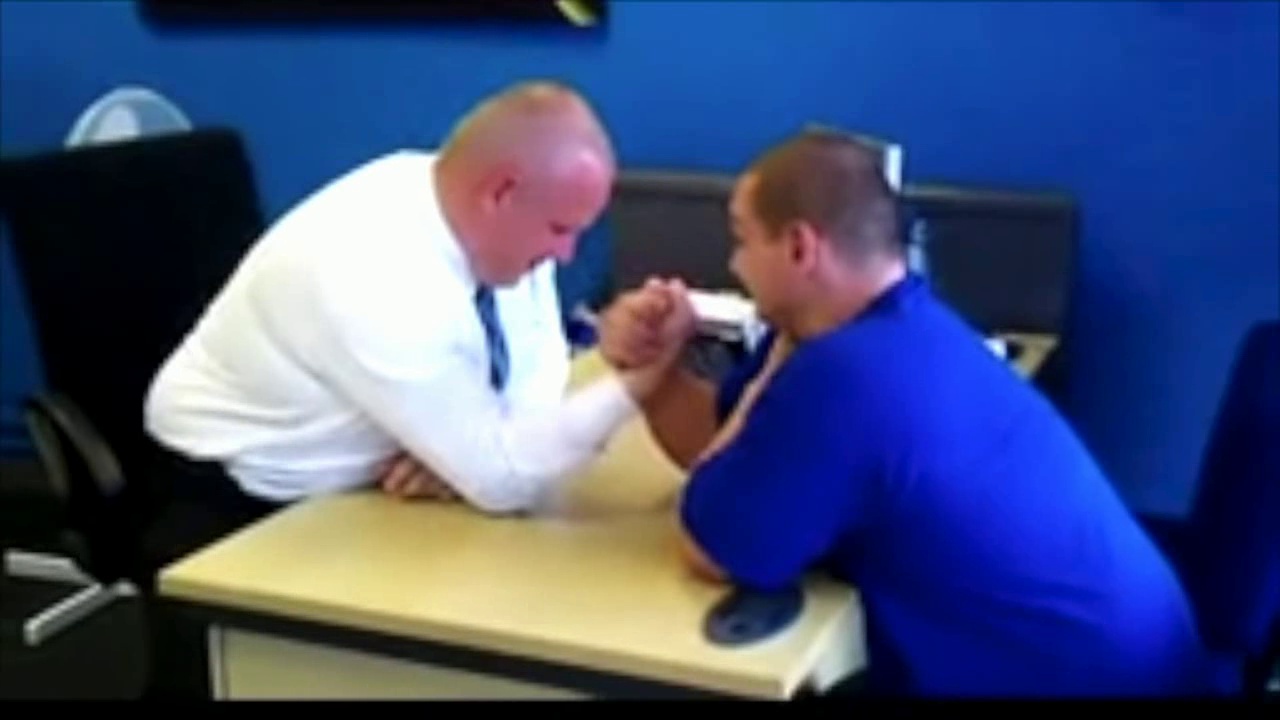}
  \includegraphics[height=\ThumbH,width=\linewidth,keepaspectratio]{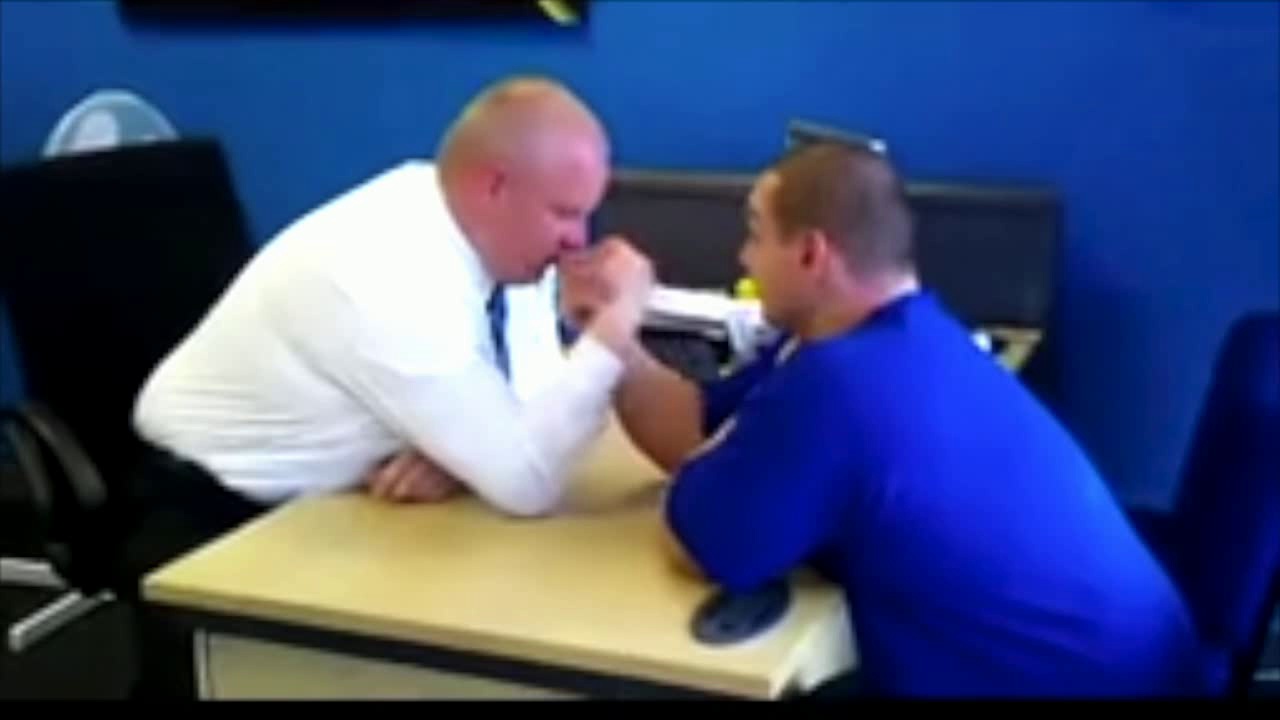}
  \includegraphics[height=\ThumbH,width=\linewidth,keepaspectratio]{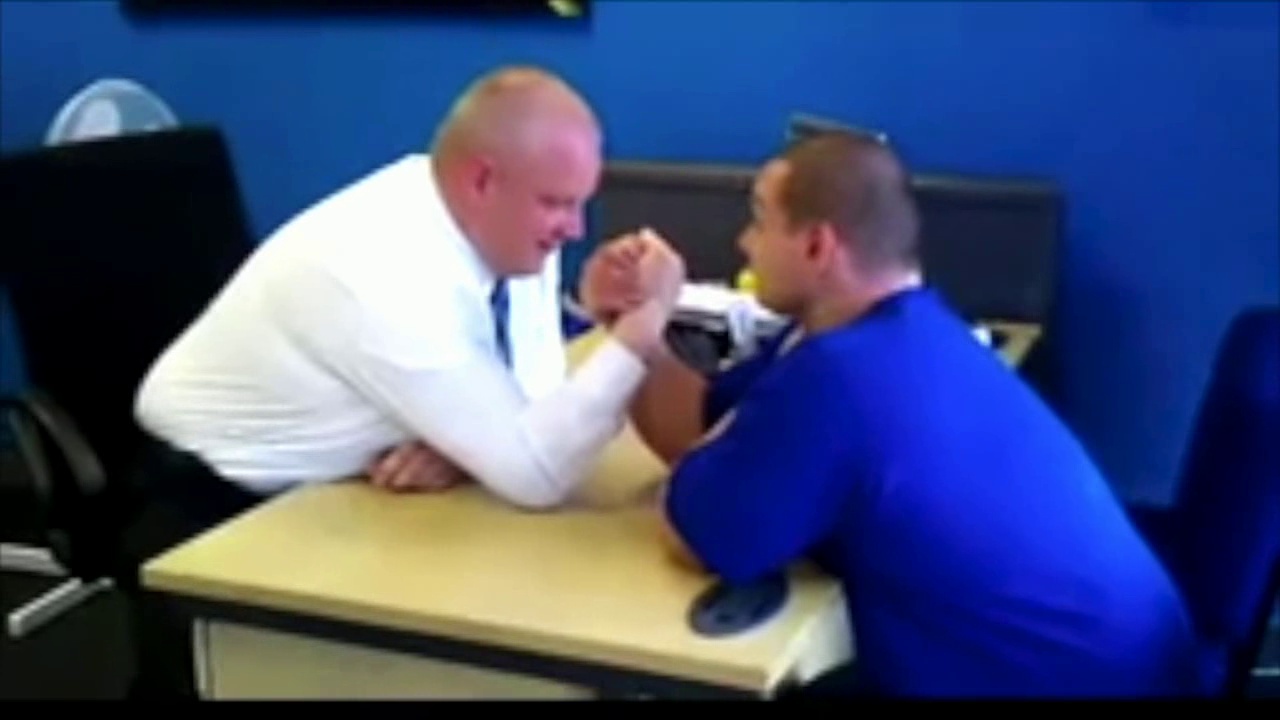}
\end{minipage}\hspace{\GapW}%
\begin{minipage}[t][\PromptTotal][t]{\dimexpr\linewidth-\LeftW-\GapW\relax}
  \vspace{0pt} % align to top
  \usebox{\PromptBoxA}\vspace{\GapW-2mm}
  \usebox{\PromptBoxB}
\end{minipage}

\vspace{2mm}

\caption{\textbf{Left}: a backward video clip (category: Put). \textbf{Top-Right}: Qwen2.5VL-72B Multi-step CoT reasoning. \textbf{Bottom-Right}: Gemini-2.5-Pro's self-explained reasoning trace. Qwen2.5VL-72B {\color{darkgreen}correctly identified the event in the scene and made a valid assumption}, but {\color{orange} failed to observe that the event was reversed} which {\color{darkred} led to an incorrect conclusion}. In contrast, Gemini-2.5-Pro {\color{darkgreen} correctly detected the reversal of the event in Step~2 based on a valid assumption it made in Step~1}. The Gemini-2.5-Pro output is from the low reasoning effort setting. }
\label{fig:combined_armwrestling_reasoning}
\end{figure*}

%---------------------------------------------
\subsection{Reasoning Effort Ablation}\label{sec:reasoning}

%\subsection{Reasoning Models vs. CoT Performance}
%Reasoning-centric models are trained to generate internal reasoning chains automatically before producing outputs, while non-reasoning models usually respond directly unless explicitly prompted for step-by-step reasoning. To control reasoning depth, we tuned the reasoning effort parameter for the former and manually designed CoT prompts for the latter.
%For proprietary reasoning models, we controlled reasoning length via the reasoning effort parameter; for non-reasoning models, we manually constructed CoT traces.
%We controlled the length of reasoning either by adjusting the reasoning effort parameter for several proprietary reasoning models or by manually constructing chain-of-thought (CoT) reasoning traces for a non-reasoning model.
\textbf{Controlling reasoning effort in proprietary reasoning VLMs.}
We varied the reasoning effort parameter (low/medium/high) for GPT-5 and Gemini-2.5-Pro (Table~\ref{tab:reasoning_results}). Surprisingly, increasing reasoning effort does not improve performance and often degrades it. Gemini-2.5-Pro with low effort achieves 61.1\% accuracy, the highest performance across all our experimental settings, yet medium and high efforts decrease to 59.9\% and 58.7\%, with Backward F1 declining from 52.1 to 51.4 to 49.0. From the low effort setting, we highlight one correctly predicted example, where we asked Gemini-2.5-Pro to explain its reasoning in a follow-up dialogue turn (bottom-right of Figure~\ref{fig:combined_armwrestling_reasoning}).  GPT-5 showed an even more pronounced pattern: low effort achieves 57.8\% accuracy, while medium and high efforts drop to 56.1\% and 56.4\%. More critically, Backward F1 collapses with increased effort (34.4 to 26.8 to 25.7), indicating amplification of forward bias. These results suggest that extended reasoning does not elicit better temporal reasoning and instead reinforces existing directional biases.

\textbf{Chain-of-thought (CoT) prompting in an open VLM.}\label{sec:cot}
We tested explicit step-by-step reasoning with Qwen2.5-VL-72B using two CoT variants (Simple and Multi-step; Section~\ref{sec:cot_settings}). As shown in Table~\ref{tab:cot_results}, CoT prompting fails to improve performance and amplifies label prediction bias, most notably with Multi-step CoT (2-shot), where Backward F1 drops by 13.2 points (Section~\ref{sec:label_prediction_bias}). In the model output example using the Multi-step CoT prompt (top-right in Figure~\ref{fig:combined_armwrestling_reasoning}), we saw that the model usually could identify the event in the video, yet it strongly believed that no physical law was violated and the video was played forward. Combined with the reasoning effort results, these findings demonstrate that additional deliberation, whether through explicit effort controls or prompted CoT, does not compensate for the lack of robust temporal and physical understanding, and often exacerbates existing biases.

\begin{table}[!tbp]
\centering
\begin{tabular}{@{}lccc@{}}
\toprule
\textbf{Data Size} & \textbf{F. F1} & \textbf{B. F1} & \textbf{Acc.} \\
\midrule

% \rowcolor{gray!15}\multicolumn{5}{c}{\textbf{Baselines}} \\
% Random & — & — & — & 50.0 \\
% Human & 30 & 99.0 & 89.0 & 89.2 \\

\rowcolor{gray!15}\multicolumn{4}{c}{\textbf{Qwen2VL-7B Training}} \\
Vanilla & 66.7 & 0.0 & 50.0 \\
1000 & 64.1 & 12.2 & 49.1 \\
3000 & 54.2 & 48.2 & 51.4 \\
5000 & 59.4 & 31.6 & 49.1 \\

\rowcolor{gray!15}\multicolumn{4}{c}{\textbf{Qwen2.5VL-7B Training}} \\
Vanilla & 63.0 & 19.5 & 49.3 \\
1000 & 54.8 & 43.5 & 49.8 \\
3000 & 54.5  & 43.4   & 49.5 \\
5000 &  54.5  &  43.4 & 49.5  \\

\bottomrule
\end{tabular}
\caption{SFT performance.}
\vspace{-3mm}
\label{tab:sft_results}
\end{table}

\begin{table}[!tbp]
\centering 
\begin{tabular}{@{}l c c c c@{}}
\toprule
\textbf{Model} & \textbf{FPS} & \textbf{F. F1} & \textbf{B. F1} & \textbf{Acc.} \\
\midrule

% \rowcolor{gray!15}\multicolumn{6}{c}{\textbf{Baselines}} \\
% \addlinespace[2pt]
% Random & — & — & — & — & 50.0 \\
% Human & — & 30 & 99.0 & 89.0 & 89.2 \\

%\rowcolor{gray!15}\multicolumn{5}{c}{\textbf{Open Models}} \\
%\addlinespace[2pt]
%\multirow{4}{*}{??} 
%  & 2 & ? & ? & ? \\
%  & 4 & ? & ? & ? \\
%  & 8 & ? & ? & ? \\
%  & 30 & ? & ? & ? \\

%\rowcolor{gray!15}\multicolumn{5}{c}{\textbf{Proprietary Models}} \\
\addlinespace[2pt]
% Human & 30&99.0 & 89.0 & \textbf{89.2}  \\ 
Human & 30& 90.0 & 88.1 & \textbf{89.2} \\

\midrule
\multirow{5}{*}{GPT-4.1} 
  % checked
  % w/o c_m 
  & 2 & 62.1 & 51.2 & 57.3 \\
  & 4 & 62.5 & 57.4 & 60.1 \\
  % & 4 (Matta) & 60.3 & 49.3 & 55.5 \\
  & 8 & 62.9 & 55.8 & 59.7 \\
  & 16 & 62.4 & 53.7 & 58.5 \\
  & 30 & 64.4 & 55.9 & \textbf{60.6} \\

\bottomrule
\end{tabular}
\caption{FPS ablation results. }
\vspace{-3mm}
\label{tab:fps_ablation_results}
\end{table}

\subsection{SFT Performance} \label{sec:sft}

We fine-tuned Qwen2-VL-7B and Qwen2.5-VL-7B on 1,000, 3,000, and 5,000 training examples to test whether explicit supervision improves AoT performance (Table~\ref{tab:sft_results}). Although fine-tuning mitigated the extreme forward bias seen in vanilla models (in the 3000 example setting, Qwen2-VL-7B Backward F1: 0→48.2, Qwen2.5-VL-7B Backward F1: 19.5→43.4), it did not lead to real task competence: accuracy stayed near chance (\textasciitilde50\%) across all training set sizes. Performance plateaued after 1,000 examples, with no gains from additional data. These results suggest that AoT task might require different training approaches beyond conventional SFT.
%We conducted supervised fine-tuning on Qwen2.5-VL-7B to investigate whether VLMs can benefit from explicit training signals for the AoT task (Table~\ref{tab:sft_results}). Despite this targeted setup, the model failed to learn meaningful patterns from the training data, and performance did not improve over the baseline. These results suggest that robust AoT competence may require stronger temporal modelling or physics-grounded training signals beyond conventional SFT.

\subsection{FPS Ablation} \label{sec:fps_ablation_results}
We evaluated GPT-4.1 across frame rates from 2 to 30 FPS and compared results to human performance (Table~\ref{tab:fps_ablation_results}). Notably, at 30 FPS, corresponding to the complete video with all frames, the model achieves only 60.6\% accuracy, falling far short of human performance at 89.2\%. This 28.6 percentage point gap demonstrates that the performance deficit is not due to insufficient temporal information. Across tested frame rates, model accuracy varies minimally (57.3–60.6\%). Given the marginal improvement from 4 FPS (60.1\%) to 30 FPS and API cost considerations, we selected 4 FPS as the default for all proprietary models.

\section{Analyses} \label{sec:analysis}
\subsection{Motion Category Analysis}

\begin{figure*}[t]
    \centering
    \includegraphics[width=2\columnwidth]{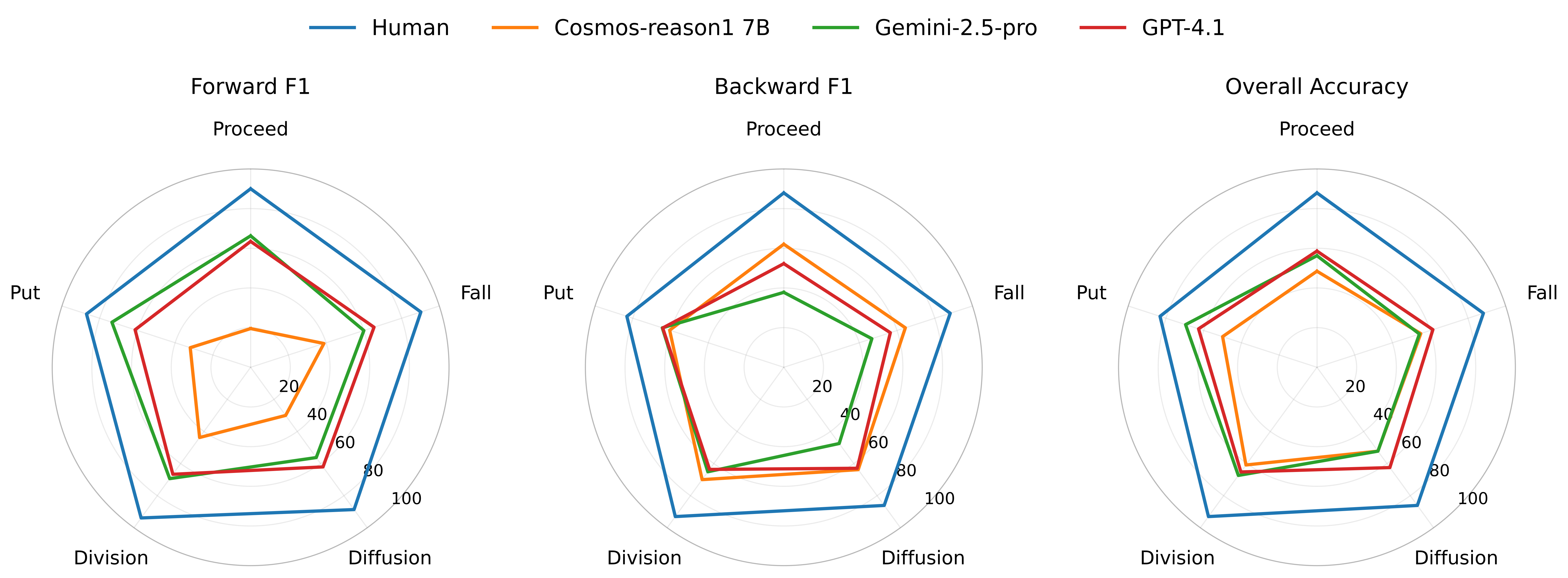}
    \caption{Per-category comparison on \benchname~ across three representative models. Cosmos-reason1-7B (zero-shot; best open-weight in this setting), GPT-4.1 (zero-shot; best proprietary and best overall model in this setting), and Gemini-2.5-Pro (zero-shot, low-reasoning effort; best model across all settings), and humans: (a) forward F1 (left), (b) backward F1 (middle), and (c) overall accuracy (right). Humans remain consistently high across all categories and both directions ($\approx$ 80-100\%). In contrast, VLMs show substantial gaps. Backward detection is the most challenging, revealing a forward-direction bias (with Cosmos-reason1 as a notable exception, showing comparatively strong backward F1). %The largest deficits appear in Forward/Proceed and Fall, with additional weaknesses in Diffusion and Addition.
    }
    \label{fig:motion_categories}
\end{figure*}

We analyze per-category performance across the five motion classes in \benchname: Proceed, Fall, Diffusion, Division, and Put. Figure~\ref{fig:motion_categories} compares humans with three VLMs: Cosmos-Reason1 7B (best open-weight model in the zero-shot setting, 52.1\%), GPT-4.1 (best proprietary and overall model in the zero-shot setting, 60.1\%), and Gemini-2.5-Pro (best overall model, 61.1\%).
Humans maintain $\approx$80-100\% accuracy across all categories and directions. VLMs show substantial gaps with striking asymmetry: relatively better on forward videos but dramatically worse on backward videos, revealing strong \textit{label prediction bias}.

\textbf{Category-specific patterns.}
The largest F1 deficits appear in Proceed and Fall, where all models struggle to capture the causal and gravitational cues defining temporal direction. Diffusion and Put show moderate F1 scores but remain below human performance even in forward playback, suggesting incomplete modeling of dispersion and goal-directed manipulation. Division is comparatively easier for both humans and models, likely due to clear temporal discontinuities (e.g., splitting). Notably, Cosmos-Reason1-7B achieves relatively strong backward F1, particularly on Fall and Division, consistent with its explicit AoT-oriented training, whereas Gemini-2.5-Pro and GPT-4.1 perform better in forward playback but drop sharply when the temporal order is reversed.
Overall, these results suggest that current VLMs rely primarily on directional visual priors and dataset correlations rather than robust physical causality understanding.

\subsection{Label Prediction Bias }
\label{sec:label_prediction_bias}
We observed that most VLMs exhibit \textbf{label prediction bias}: models strongly favor one label over the other despite balanced forward/backward video distribution in \benchname. In zero-shot evaluation (Table~\ref{tab:zero_shot_results}), GPT-4o predicted 87\% of clips as Forward (only 13\% as Backward), while QVQ-72B-Preview predicted 100\% Forward, causing minority-class F1 scores to drop below 0.4 due to low recall. Most models exhibit Forward bias (Qwen2.5-VL series, QVQ-72B-Preview, GPT-4o, o3, o4-mini, GPT-5), while Cosmos-Reason1-7B shows Backward bias, likely due to extensive reversed-clip exposure during reinforcement learning. 

Furthermore, \textbf{reasoning amplifies bias}: when models engage in step-by-step reasoning, biases intensify rather than improve. QVQ-72B-Preview often correctly identified scene events but failed to detect reversed motion, with its reasoning process reinforcing the idea that the video is played forward. Similarly, increasing reasoning effort (GPT-5 in Table~\ref{tab:reasoning_results}) or enforcing chain-of-thought (Table~\ref{tab:cot_results}, Figure~\ref{fig:combined_armwrestling_reasoning}) further amplified label prediction bias rather than correcting it. %These findings suggest that current VLMs lack robust internal models of temporal causality, with reasoning mechanisms that compound rather than correct perceptual biases.

\section{Conclusion}

We present \textbf{\benchname}, a psychophysically validated benchmark designed to assess whether modern vision–language models (VLMs) possess a human-like inductive bias for inferring the \textit{arrow of time}—the implicit understanding that physical events unfold irreversibly from past to future. Our experiments reveal that current VLMs perform far below human baselines, even on intuitive scenarios such as falling objects, which humans find easy to interpret.
This striking shortfall cannot be attributed to limited data or lack of reasoning depth. Instead, it highlights a fundamental absence of inductive biases for temporal continuity, causality, and physical irreversibility—principles that humans internalize effortlessly through interaction with the physical world. We release \benchname~ and its evaluation code to foster the development of multimodal systems that move beyond statistical pattern recognition toward genuine physical understanding.

% We present \textbf{AoT-PsyPhyBENCH}, a psychophysically validated benchmark to test if modern vision-language models possess the inductive bias like humans to infer the \emph{arrow of time}—the implicit understanding that physical events unfold irreversibly from past to future. Our results show that VLMs lag significantly behind human baseline, even on easier cases such as objects falling or division, where humans find extremely easy to distinguish. 
% This stricking failure cannot be explained by insufficient data or reasoning depth; rather it reflects the absence of inductive biases for temporal continuity, causality, and physical irreversibility, principles that human observers acquire effortlessly through lifelong interaction with the physical world. 
% We release the benchmark and evaluation code to enable future development of multimodal systems, in hope of bridging the gap between statistical pattern recognition and genuinely physical understanding.

% 1. we present bench.
% 2. we provide results of models, and compare with human, categorized
% 3. we notice gaps and bias. 
% 4. we release the benchmark and the code. 

\section{Limitations}
Our work has two limitations. % First, we only tested some of the most recent VLMs that we judged representative at the time to the . 
%First, for proprietary models, even though we report the specific model names and versions used, exact replication may be affected by undisclosed internal updates or serving-time changes beyond our control. 
First, for proprietary models, even though we report the specific model names and versions used, exact replication may be affected by undisclosed updates or API changes beyond our control. 
Second, we haven't been able to offer a conclusive explanation for why SFT did not work on AoT. One possible reason is that open-source VLMs may not effectively encode temporal information along the frame sequence. Cosmos-Reason1-7B saw improvement with reinforcement learning in their work, however, they did not publish their dataset and we did not see improvement on our benchmark, either. 

% We haven't been able to offer a conclusive explanation for why SFT did not work on AOT. One possible reason is that open-source VLM may not effectively encode temporal information along the frame sequence. Cosmos-Reason1-7B saw improvement with reinforcement learning in their work, however, they did not publish their dataset and we did not see improvement on our benchmark, either. 

% Cosmos-Reason1-7B

% First, task scope is narrow since clips are only about three seconds and the decision is a forward or backward label, and we exclude cyclic motions, so ecological validity is limited and some temporal cues are under tested. Second, we used low frame rates due to cost and model constraints, but certain phenomena may need denser sampling to expose subtle motion cues. Third, although the benchmark is tied to human baselines, a perfectly matched comparison is not possible because participant viewing conditions and response style differ from the models inputs and outputs, so scores should be read as approximate rather than iso-informational.

%\acknowledgments
\bibliography{custom}

@article{agarwal2025cosmos,
  title={Cosmos world foundation model platform for physical ai},
  author={Agarwal, Niket and Ali, Arslan and Bala, Maciej and Balaji, Yogesh and Barker, Erik and Cai, Tiffany and Chattopadhyay, Prithvijit and Chen, Yongxin and Cui, Yin and Ding, Yifan and others},
  journal={arXiv preprint arXiv:2501.03575},
  year={2025}
}

@article{azzolini2025cosmos,
  title={Cosmos-reason1: From physical common sense to embodied reasoning},
  author={Azzolini, Alisson and Bai, Junjie and Brandon, Hannah and Cao, Jiaxin and Chattopadhyay, Prithvijit and Chen, Huayu and Chu, Jinju and Cui, Yin and Diamond, Jenna and Ding, Yifan and others},
  journal={arXiv preprint arXiv:2503.15558},
  year={2025}
}

@misc{openai_gpt4o_blog_2024,
  author       = {OpenAI},
  title        = {Hello GPT-4o},
  year         = {2024},
  month        = may,
  howpublished = {\url{https://openai.com/index/hello-gpt-4o/}}
}

@misc{openai_gpt41_blog_2025,
  author = {OpenAI},
  title = {Introducing GPT-4.1 in the API},
  url = {https://openai.com/index/gpt-4-1/},
  year = {2025},
  organization = {Openai.com}
}

@misc{openai_gpt5_blog_2025,
  author = {OpenAI},
  title = {Introducing GPT-5},
  url = {https://openai.com/index/introducing-gpt-5/},
  year = {2025},
  organization = {Openai.com}
}

@misc{openai_o3_blog_2025,
  author       = {OpenAI},
  title        = {Introducing OpenAI o3 and o4-mini},
  year         = {2025},
  month        = apr,
  howpublished = {\url{https://openai.com/index/introducing-o3-and-o4-mini/}}
}

@techreport{qwen_qvq_2024,
  title={QVQ: Qwen2-VL with Visual Question Answering},
  author={{Qwen Team}},
  year={2024},
  institution={Alibaba Cloud},
  url={https://qwenlm.github.io/blog/qvq-72b-preview/}
}

@article{wang2024qwen2vl,
  title={Qwen2-VL: Enhancing Vision-Language Model's Perception of the World at Any Resolution},
  author={Wang, Peng and Bai, Shuai and Tan, Sinan and Wang, Shijie and Fan, Zhihao and Bai, Jinze and Chen, Keqin and Liu, Xuejing and Wang, Jialin and Ge, Wenbin and Fan, Yang and Dang, Kai and Du, Mengfei and Ren, Xuancheng and Men, Rui and Liu, Dayiheng and Zhou, Chang and Zhou, Jingren and Lin, Junyang},
  journal={arXiv preprint arXiv:2409.12191},
  year={2024},
  url={https://arxiv.org/abs/2409.12191}
}

@article{bai2025qwen2,
  title={Qwen2. 5-vl technical report},
  author={Bai, Shuai and Chen, Keqin and Liu, Xuejing and Wang, Jialin and Ge, Wenbin and Song, Sibo and Dang, Kai and Wang, Peng and Wang, Shijie and Tang, Jun and others},
  journal={arXiv preprint arXiv:2502.13923},
  year={2025}
}

@article{monfort2019moments,
  title={Moments in time dataset: one million videos for event understanding},
  author={Monfort, Mathew and Andonian, Alex and Zhou, Bolei and Ramakrishnan, Kandan and Bargal, Sarah Adel and Yan, Tom and Brown, Lisa and Fan, Quanfu and Gutfreund, Dan and Vondrick, Carl and others},
  journal={IEEE transactions on pattern analysis and machine intelligence},
  volume={42},
  number={2},
  pages={502--508},
  year={2019},
  publisher={IEEE}
}

@article{comanici2025gemini,
  title={Gemini 2.5: Pushing the frontier with advanced reasoning, multimodality, long context, and next generation agentic capabilities},
  author={Comanici, Gheorghe and Bieber, Eric and Schaekermann, Mike and Pasupat, Ice and Sachdeva, Noveen and Dhillon, Inderjit and Blistein, Marcel and Ram, Ori and Zhang, Dan and Rosen, Evan and others},
  journal={arXiv preprint arXiv:2507.06261},
  year={2025}
}

@article{xue2025seeing,
  title={Seeing the Arrow of Time in Large Multimodal Models},
  author={Xue, Zihui and Luo, Mi and Grauman, Kristen},
  journal={arXiv preprint arXiv:2506.03340},
  year={2025}
}

@inproceedings{bagad2023test,
  title={Test of time: Instilling video-language models with a sense of time},
  author={Bagad, Piyush and Tapaswi, Makarand and Snoek, Cees GM},
  booktitle={Proceedings of the IEEE/CVF Conference on Computer Vision and Pattern Recognition},
  pages={2503--2516},
  year={2023}
}

@inproceedings{du2024reversed,
  title={Reversed in time: A novel temporal-emphasized benchmark for cross-modal video-text retrieval},
  author={Du, Yang and Liu, Yuqi and Jin, Qin},
  booktitle={Proceedings of the 32nd ACM International Conference on Multimedia},
  pages={5260--5269},
  year={2024}
}

@article{wang2023paxion,
  title={Paxion: Patching action knowledge in video-language foundation models},
  author={Wang, Zhenhailong and Blume, Ansel and Li, Sha and Liu, Genglin and Cho, Jaemin and Tang, Zineng and Bansal, Mohit and Ji, Heng},
  journal={Advances in Neural Information Processing Systems},
  volume={36},
  pages={20729--20749},
  year={2023}
}

@article{hu2021lora,
  title   = {LoRA: Low-Rank Adaptation of Large Language Models},
  author  = {Hu, Edward J. and Shen, Yelong and Wallis, Phillip and Allen-Zhu, Zeyuan and Li, Yuanzhi and Wang, Lu and Chen, Weizhu},
  journal = {arXiv preprint arXiv:2106.09685},
  year    = {2021},
  url     = {https://arxiv.org/abs/2106.09685}
}

@article{hanyu2023ready,
  title={Ready to detect a reversal of time's arrow: a psychophysical study using short video clips in daily scenes},
  author={Hanyu, Nao and Watanabe, Kei and Kitazawa, Shigeru},
  journal={Royal Society open science},
  volume={10},
  number={4},
  pages={230036},
  year={2023},
  publisher={The Royal Society}
}

@inproceedings{liu-etal-2024-mibench,
    title = "{MIB}ench: Evaluating Multimodal Large Language Models over Multiple Images",
    author = "Liu, Haowei  and
      Zhang, Xi  and
      Xu, Haiyang  and
      Shi, Yaya  and
      Jiang, Chaoya  and
      Yan, Ming  and
      Zhang, Ji  and
      Huang, Fei  and
      Yuan, Chunfeng  and
      Li, Bing  and
      Hu, Weiming",
    editor = "Al-Onaizan, Yaser  and
      Bansal, Mohit  and
      Chen, Yun-Nung",
    booktitle = "Proceedings of the 2024 Conference on Empirical Methods in Natural Language Processing",
    month = nov,
    year = "2024",
    address = "Miami, Florida, USA",
    publisher = "Association for Computational Linguistics",
    url = "https://aclanthology.org/2024.emnlp-main.1250/",
    doi = "10.18653/v1/2024.emnlp-main.1250",
    pages = "22417--22428"
}

@misc{wang2024muirbench,
      title={MuirBench: A Comprehensive Benchmark for Robust Multi-image Understanding}, 
      author={Fei Wang and Xingyu Fu and James Y. Huang and Zekun Li and Qin Liu and Xiaogeng Liu and Mingyu Derek Ma and Nan Xu and Wenxuan Zhou and Kai Zhang and Tianyi Lorena Yan and Wenjie Jacky Mo and Hsiang-Hui Liu and Pan Lu and Chunyuan Li and Chaowei Xiao and Kai-Wei Chang and Dan Roth and Sheng Zhang and Hoifung Poon and Muhao Chen},
      year={2024},
      eprint={2406.09411},
      archivePrefix={arXiv},
      primaryClass={cs.CV},
      url={https://arxiv.org/abs/2406.09411}, 
}

@misc{fu2025videomme,
      title={Video-MME: The First-Ever Comprehensive Evaluation Benchmark of Multi-modal LLMs in Video Analysis}, 
      author={Chaoyou Fu and Yuhan Dai and Yongdong Luo and Lei Li and Shuhuai Ren and Renrui Zhang and Zihan Wang and Chenyu Zhou and Yunhang Shen and Mengdan Zhang and Peixian Chen and Yanwei Li and Shaohui Lin and Sirui Zhao and Ke Li and Tong Xu and Xiawu Zheng and Enhong Chen and Caifeng Shan and Ran He and Xing Sun},
      year={2025},
      eprint={2405.21075},
      archivePrefix={arXiv},
      primaryClass={cs.CV},
      url={https://arxiv.org/abs/2405.21075}, 
}

@misc{xiao2021nextqa,
      title={NExT-QA:Next Phase of Question-Answering to Explaining Temporal Actions}, 
      author={Junbin Xiao and Xindi Shang and Angela Yao and Tat-Seng Chua},
      year={2021},
      eprint={2105.08276},
      archivePrefix={arXiv},
      primaryClass={cs.CV},
      url={https://arxiv.org/abs/2105.08276}, 
}

@misc{wu2024starbench,
      title={STAR: A Benchmark for Situated Reasoning in Real-World Videos}, 
      author={Bo Wu and Shoubin Yu and Zhenfang Chen and Joshua B Tenenbaum and Chuang Gan},
      year={2024},
      eprint={2405.09711},
      archivePrefix={arXiv},
      primaryClass={cs.AI},
      url={https://arxiv.org/abs/2405.09711}, 
}

\appendix
%\section{Example Appendix}
%\label{sec:appendix}
%This is an appendix.
\section{Appendix}

\subsection{Instructions in Manual CoT Experiments}

The instruction used in the Simple CoT setting is shown in Figure~\ref{fig:simple_cot_instruction}. The instruction for the Multi-step CoT setting is shown in Figure~\ref{fig:multi_step_cot_instruction}.

\begin{figure}[!h]
\centering
% top box
\begin{minipage}{1\linewidth}
\begin{lstlisting}[style=promptbox]
(<<bf>>Instruction<</bf>>) Is this video played forward (F) or backward (B)? Pay attention to moving parts in the video as they may be the visual clues to indicate the direction of the playback. Think step by step before giving out your final answer. Finish your answer with F for forward or B for backward only.
\end{lstlisting}
\end{minipage}

% % bottom box
% \begin{minipage}{\textwidth}
% \begin{lstlisting}[style=promptbox]
% (<<bf>>Hand-curated Simple CoT Output<</bf>>) This scene appears to take place on the water, where a large splash is occurring. It looks like an explosion happened underwater. The splash is very large at the beginning but gradually becomes smaller and eventually converges to a single point. (......) Since the splash starts out large and gradually converges to a small point, it contradicts the physical law that an explosion should begin small and expand outward. Therefore, this video is being played backward. <<bf>>B<</bf>>
% \end{lstlisting}
% \end{minipage}

\caption{The instruction for the Simple CoT setting.}
\label{fig:simple_cot_instruction}
\end{figure}

\begin{figure}[!h]
\centering
\begin{minipage}{1\linewidth}
\begin{lstlisting}[style=promptbox]
Is this video played forward or backward? 
Think in three steps: (*@\textbf{Observation}@*), (*@\textbf{Assumption}@*), and (*@\textbf{Conclusion}@*). 
In (*@\textbf{Observation}@*), describe what you see in the video honestly, without making any early assumptions about whether it is played forward or backward. Pay close attention to moving elements such as people's movements, the flow of liquid, or any changes in the size or position of objects. In (*@\textbf{Assumption}@*), make a reasonable guess about what should normally happen in this scene if it were played forward. Base this guess on common sense, including causal relationships, the laws of physics, and the arrow of time. In (*@\textbf{Conclusion}@*), check whether your Observation matches or contradicts your Assumption. If what you observed follows the expected behavior based on your assumption, then the video is played forward. If what you observed goes against your assumption, then the video is being played backward.
Finish your answer with F for forward or B for backward only.
\end{lstlisting}
\end{minipage}
\caption{The instruction used for the Multi-step CoT setting.}
\label{fig:multi_step_cot_instruction}
\end{figure}

\end{document}